\documentclass[ASNA,twocolumn]{USG}

\usepackage{anyfontsize}
\usepackage{amsmath,amsfonts}
\usepackage{array}
\usepackage{booktabs}
\usepackage{siunitx}
\usepackage{tikz}
\usetikzlibrary{arrows.meta, patterns, calc, positioning}

\graphicspath{{./Figures/}{./images/}}

\articletype{RESEARCH ARTICLE}

\received{00 Month 0000}
\revised{00 Month 0000}
\accepted{00 Month 0000}
\journal{Journal of Field Robotics}
\volume{00}
\copyyear{0000}
\startpage{1}
\articledoi{10.1002/0000}

\begin{document}


\title{Design and Experimental Validation of the Separation and Deployment System for the CHERI Lunar Micro-Rovers}

\author[1]{Ahmet Mustafa Kangal}
\author[1]{O\u{g}uzhan \c{S}ahan}  
\author[2]{Teoman Efe Aycan} 
\author[2]{G\"{o}rkem Egemen G\"{u}lo\u{g}lu}
\author[2]{Halil Ersin S\"{o}ken}
\author[1]{Ali Emre Turgut}
\author[1]{Uluç Saranl{\i}}
\author[1]{Mustafa Mert Ankaral{\i}}

\address[1]{\orgdiv{Robotics Graduate Program, Graduate School of Natural and Applied Sciences, }\orgname{Middle East Technical University (METU), }%
\orgaddress{\state{Ankara, }\country{T\"{u}rkiye}}}

\address[2]{\orgdiv{Department of Aerospace Engineering, }\orgname{Middle East Technical University (METU), }%
\orgaddress{\state{Ankara, }\country{T\"{u}rkiye}}}

\authormark{KANGAL \textsc{et al.}}
\titlemark{SEPARATION AND DEPLOYMENT SYSTEM FOR THE CHERI LUNAR MICRO-ROVERS}

\corres{Ahmet Mustafa Kangal (\email{ahmet.kangal@metu.edu.tr})}


\abstract[ABSTRACT]{Reliable separation mechanisms are critical for planetary micro-rovers, which must withstand launch vibrations and then eject the vehicle clear of the lander. This paper presents the design, optimization, and testing of the separation system of the CHERI (Challenging Environment Exploration Rovers for Intelligence) lunar micro-rovers. Sized for a $5\,\text{kg}$ vehicle, the assembly comprises an aluminum ground plate, a central non-explosive hold-down and release mechanism, and an asymmetric four-spring ejection array. Initial separation trials revealed a velocity deficit caused by energy absorption in the flexible wheel-legs and chassis during the ejection stroke. This loss was represented as equivalent chassis damping in a multi-body dynamic model, and the springs were re-optimized accordingly. Early hardware iterations failed vibration testing, first through spring jamming in the guide housings and then through a $34\,\text{Hz}$ fundamental mode of the flat ground plate. A density-based topology optimization raised the fundamental frequency to $84\,\text{Hz}$, above the $70\,\text{Hz}$ requirement of the lander, within a $1.6\,\text{kg}$ mass budget. The final configuration passed acceptance-level sinusoidal vibration, random vibration, and shock testing without structural degradation. Separation tests with a weight-matched one-sixth mass model, tracked by optical motion capture, confirmed clearance of the lander and reproduced the simulated free flight to within the run-to-run variation of the measured ejection velocity. These tests further showed that landing outcome is dominated by touchdown and settling dynamics, indicating that the residual pitch rate is the main controllable design parameter and confirming self-righting recovery as an integral element of the deployment architecture.}

\maketitle

\section{Introduction}
\label{sec:introduction}

The global surge in lunar exploration, particularly targeting the volatile-rich lunar South Pole, has renewed interest in complementing heavy, monolithic exploration platforms with coordinated networks of micro-rovers~\cite{ellery2016planetary, schweitzer2022micro}. These ultra-lightweight vehicles (typically $ \leq 10\,\text{kg}$) function as cost-effective, high-yield scouts capable of approaching high-risk terrains, such as the vicinity of permanently shadowed regions (PSRs), where parent spacecraft cannot safely land. Operating as cooperative multi-agent networks, micro-rovers can perform distributed scientific measurements, localized mapping, and autonomous heterogeneous tasks, increasing the scientific return per launch mass. However, the extreme thermal environment (surface temperatures above $+100^\circ\text{C}$ in sunlight and below $-230^\circ\text{C}$ inside PSRs), abrasive lunar dust, and high vacuum of the lunar South Pole~\cite{heiken1991lunar} place severe engineering constraints on the physical co-design of their structural, mobility, and deployment subsystems.

A critical engineering bottleneck in micro-rover missions is the physical separation and deployment of the vehicle from the host lander. While conventional heavy-class rovers utilize complex, heavy motorized ramps or articulated robotic arms to safely reach the regolith, micro-rovers cannot tolerate such substantial mass penalties. Consequently, passive ``drop-deployment'' strategies, in which the vehicle is released from its hold-down and falls or is spring-ejected directly onto the terrain, have emerged as a mass-efficient alternative~\cite{schweitzer2022micro}. However, free-fall deployment from high mounting points introduces significant risk: the vehicle must clear the lander's footpads to avoid physical collision while maintaining rotational stability in flight to prevent tumbling or tipping over upon landing. Designing a mechanism that ensures clean kinetic separation within a highly restricted mass allocation is therefore a multi-disciplinary optimization problem. To mitigate the risk of landing in an inverted state, the deployment strategy relies on two layers: the separation system is designed to control the rover's attitude in flight and reduce the probability of an inverted touchdown, and the rover incorporates a self-righting recovery mechanism to restore an upright attitude when the first layer is not sufficient.

\begin{figure}[htbp]
\centering
\includegraphics[width=\columnwidth]{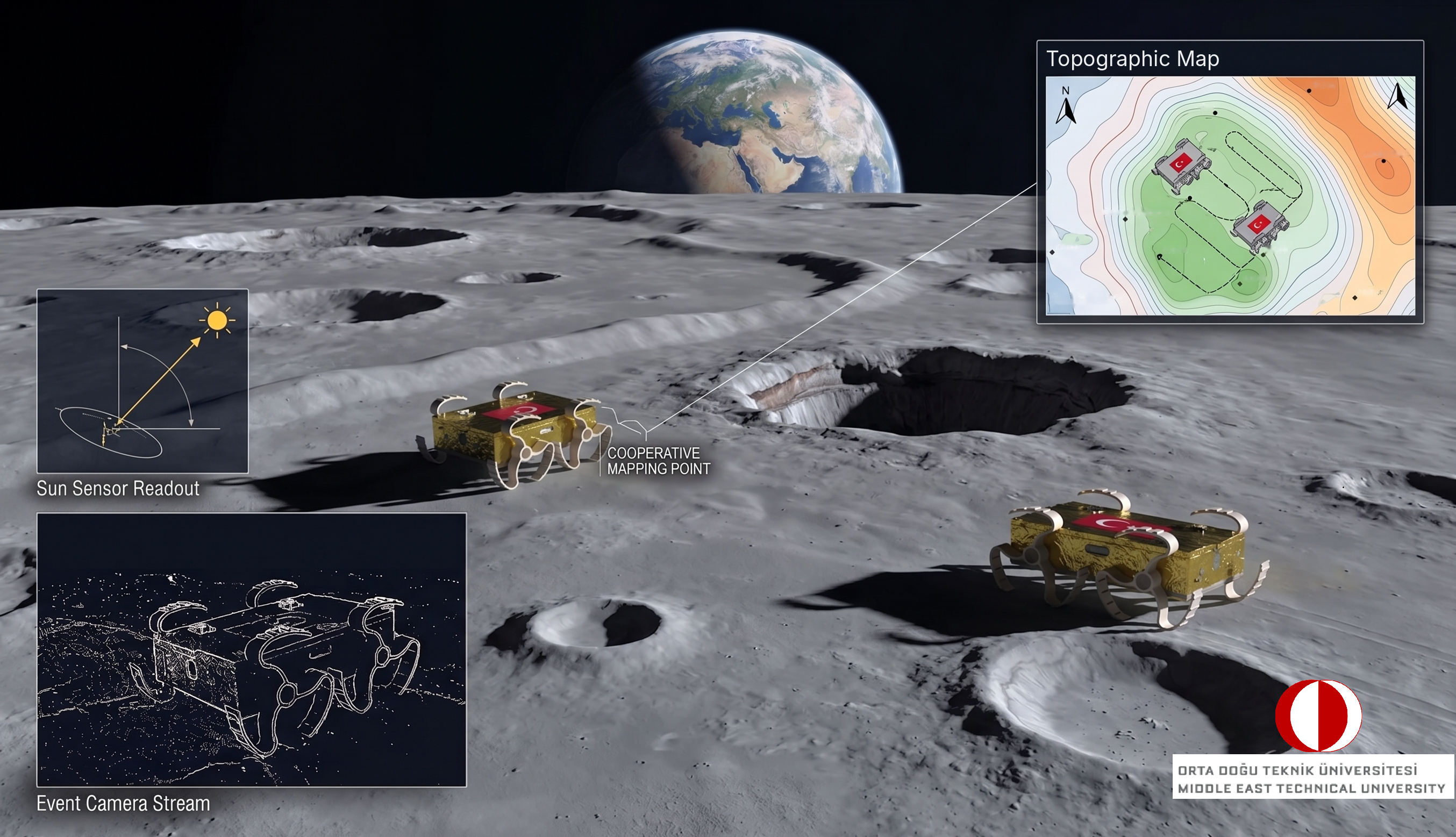}
\caption{Concept illustration of the CHERI micro-rovers (AI-generated image, shown for illustration only).}
\label{fig:concept}
\end{figure}

The Challenging Environment Exploration Rovers for Intelligence (CHERI) project, developed at Middle East Technical University (METU), T\"urkiye, addresses these design challenges. Engineered as a pair of lightweight ($5.0\,\text{kg}$ each) micro-rovers, the CHERI platform is designed to demonstrate multi-robot autonomous coordination, cooperative mapping, and surface mobility, and sized within the payload interface limits of a host lunar lander~\cite{Ankarali2026_design}, as illustrated in Fig. \ref{fig:concept}. To deploy these vehicles without heavy auxiliary ramps, the rovers are mounted laterally on the lander body at a height of $H = 1.7\,\text{m}$ above the regolith and must perform a passive, spring-actuated free-fall landing. To ensure safe deployment, the separation mechanism's primary objective is to propel each vehicle laterally by at least $D_{\text{clearance}} \ge 0.7\,\text{m}$ to clear the lander's primary structural footpads, while managing in-flight kinematics to promote a flat touchdown attitude. Because in-flight attitude control alone cannot guarantee an upright landing on every terrain condition, CHERI integrates a secondary, active recovery system as an essential layer rather than a rarely-invoked contingency: if the vehicle lands upside down, coordinated leg maneuvers combined with the deployment of its single, top-mounted solar panel act as physical levers to push against the lunar regolith and self-right the chassis to restore nominal driving mobility.

\begin{figure}[tb]
\centering

\begin{minipage}{0.48\columnwidth}
    \centering
    \includegraphics[width=\linewidth]{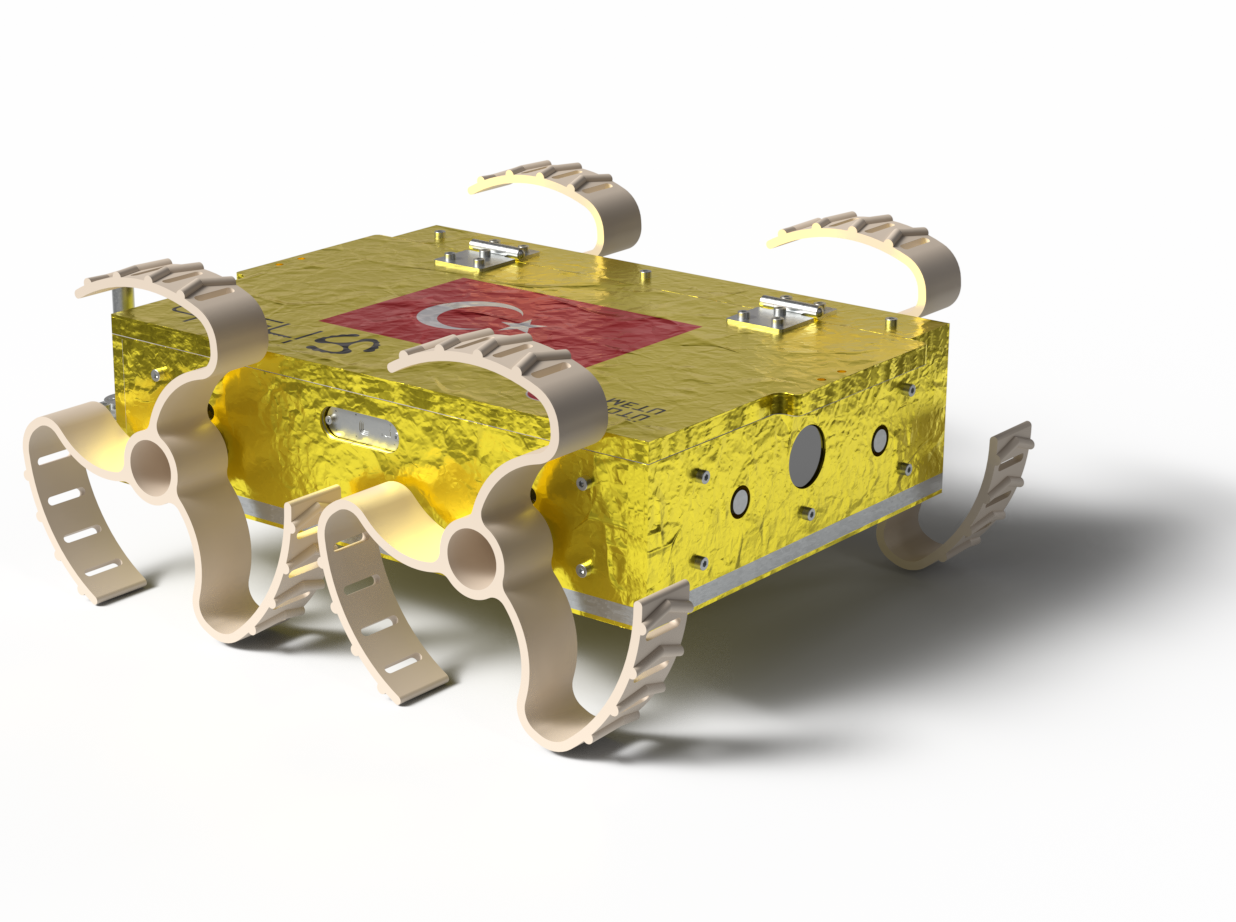}\vspace{1mm}\\
    \small (a) CAD concept render
\end{minipage}\hfill
\begin{minipage}{0.48\columnwidth}
    \centering
    \includegraphics[width=\linewidth]{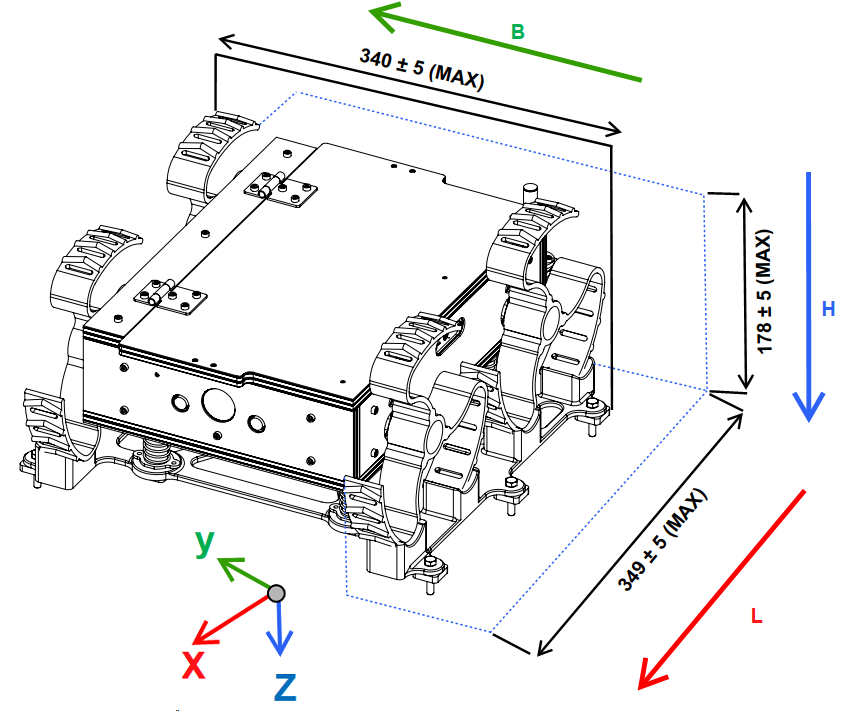}\vspace{1mm}\\
    \small (b) Technical layout
\end{minipage}

\vspace{4mm} 

\begin{minipage}{\columnwidth}
    \centering
    \includegraphics[width=0.9\linewidth]{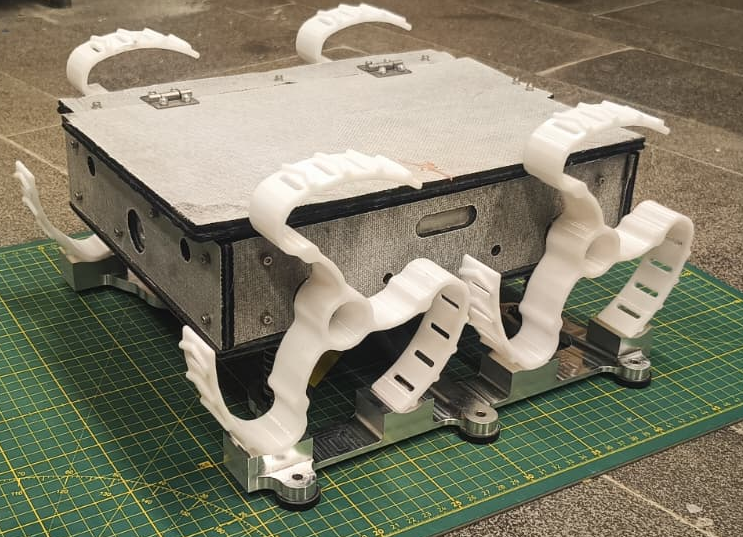}\vspace{1mm}\\
    \small (c) Manufactured Structural Model (SM)
\end{minipage}

\caption{The CHERI micro-rover platform: (a) CAD render of the rover; (b) orthographic technical drawing indicating key structural dimensions and the hold-down interface; (c) the manufactured Structural Model (SM) used for mechanical verification and environmental testing.}
\label{fig:cheri_system_architecture}
\end{figure}

\begin{figure}[htbp]
\centering
\begin{tikzpicture}[scale=1.4, >=Stealth, 
    lander/.style={draw, thick, fill=gray!20},
    rover/.style={draw, thick, fill=blue!10},
    ground/.style={draw, thick, fill=orange!10}
]
    \draw[thick] (-0.5,0) -- (4,0);
    \foreach \x in {-0.5,-0.25,0,0.25,0.5,0.75,1.0,1.25,1.5,1.75,2.0,2.25,2.5,2.75,3.0,3.25,3.5,3.75} {
        \draw (\x,0) -- (\x-0.12,-0.12);
    }
    \node[scale=0.8] at (3.4,-0.25) {Lunar Regolith};

    \draw[lander] (0,0) -- (0.15,0.3) -- (0.35,0.3) -- (0.5,0) -- cycle; 
    \draw[thick] (0.25,0.3) -- (0.25,1.7) -- (-0.5,1.7); 
    \draw[fill=gray!40] (-0.6, 1.2) rectangle (-0.2, 2.1); 
    \node[anchor=south, scale=0.8] at (-0.4, 2.1) {Lander Body};

    \draw[dashed, draw=gray, fill=gray!10] (-0.2, 1.35) rectangle (-0.02, 1.95); 
    \draw[dashed, draw=gray, fill=gray!20] (-0.02, 1.45) circle (0.08); 
    \draw[dashed, draw=gray, fill=gray!20] (-0.02, 1.85) circle (0.08);
    \node[scale=0.6, text=gray, rotate=90] at (-0.11, 0.65) {Stowed (Vertical)};

    \draw[->, thick, dashed, red] (0.05, 1.65) parabola bend (0.6,1.72) (2.5, 0.25);
    \node[red, scale=0.7, anchor=south west] at (1.4,1.3) {Trajectory};

    \draw[rover] (2.3, 0.1) rectangle (2.9, 0.28); 
    \draw[thick, fill=black!30] (2.4, 0.1) circle (0.1); 
    \draw[thick, fill=black!30] (2.8, 0.1) circle (0.1);
    \node[scale=0.6] at (3.4, 0.19) {CHERI Rover};

    \draw[<->, thick, blue] (-0.5, 0) -- (-0.5, 1.65) node[midway, left, scale=0.8] {$H = 1.7\,\text{m}$};
    \draw[dotted] (-0.5,1.65) -- (-0.2,1.65);
    
    \draw[<->, thick, green!60!black] (0.4, 0.4) -- (2.3, 0.4) node[midway, above, scale=0.8] {$D_{\text{clearance}} \ge 0.7\,\text{m}$};
    \draw[dotted] (0.4,0) -- (0.4,0.5);
    \draw[dotted] (2.3,0) -- (2.3,0.5);

\end{tikzpicture}
\caption{Free-fall deployment kinematics showing the stowed vertical mounting height ($H = 1.7\,\text{m}$) and the minimum dynamic horizontal clearance ($D_{\text{clearance}} \ge 0.7\,\text{m}$) required to clear the host lander's landing pads.}
\label{fig:deployment_schematic}
\end{figure}
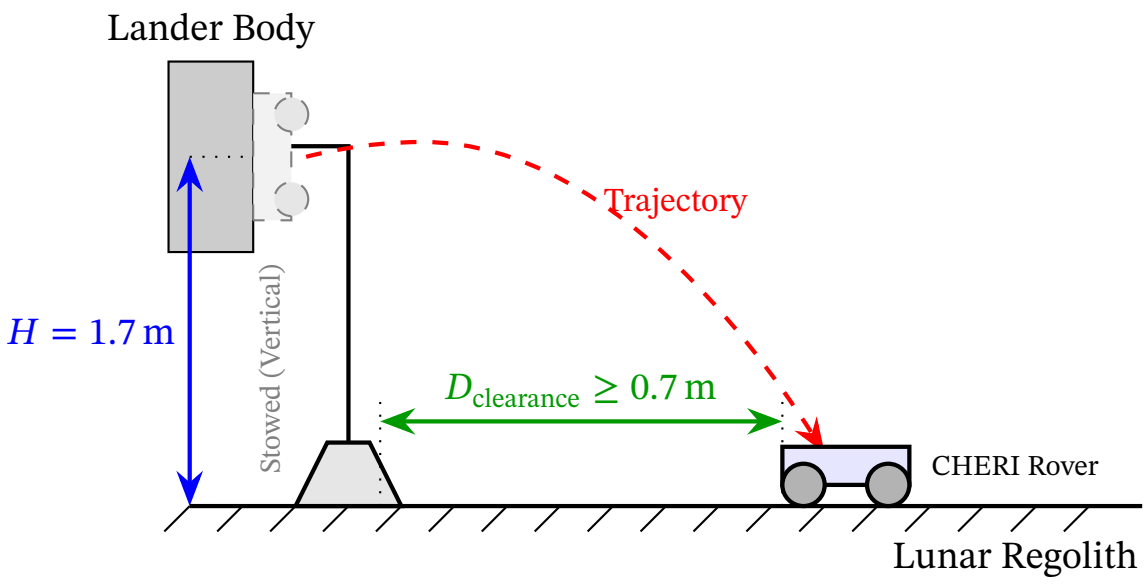

The physical design of the CHERI system and its deployment mechanisms are illustrated in Fig.~\ref{fig:cheri_system_architecture}. To show the translation from conceptual design to physical validation, Fig.~\ref{fig:cheri_system_architecture}(a) presents the CAD render of the CHERI micro-rover, showing its J-shaped flexible wheel-legs and structural layout. Fig.~\ref{fig:cheri_system_architecture}(b) provides an orthographic technical drawing detailing the primary vehicle dimensions and the bottom-mounted mechanical separation interface. Finally, Fig.~\ref{fig:cheri_system_architecture}(c) showcases the manufactured Structural Model (SM) of the rover assembly utilized in ground testing campaigns. The corresponding free-fall deployment schematic and its spatial clearance constraints are illustrated in Fig.~\ref{fig:deployment_schematic}.

While individual subsystems such as shock-absorbing lander structures and launch locks are well studied, there is little published work on an integrated design-optimization-verification methodology in the literature to guide the co-development of micro-rover physical deployment components. Spring-ejection systems are often designed under ideal rigid-body assumptions, ignoring the parasitic energy absorption caused by the structural compliance of lightweight chassis and flexible mobility limbs. This paper addresses this gap by presenting the end-to-end development of the CHERI separation system, showing how analytical sizing, multi-body dynamics, optimization, and experimental ground testing were combined, including the limits of what the ground tests could establish.

The primary contributions of this work are:
\begin{enumerate}
\item \textbf{Asymmetric Spring Array:} Design of an off-center, asymmetric four-spring ejection array relative to the vehicle's Center of Mass (CoM), which imparts a controlled pitch rate during separation to bias the touchdown attitude, together with an experimental assessment of landing outcomes over terrain slopes, which identifies the touchdown phase, rather than free flight, as the limit on baseline landing reliability.
\item \textbf{Parasitic Compliance Modeling \& Recalibration:} Identification of the energy absorbed by the J-shaped flexible wheel-legs and chassis during the separation stroke, its representation as an equivalent chassis damping in the multi-body model, and comparison of the resulting free-flight predictions with 6-DOF optical motion-capture measurements of a weight-matched $1/6$ mass model.
\item \textbf{High-Stiffness, Topology-Optimized Ground Plate:} Implementation of SIMP (Solid Isotropic Material with Penalization) topology optimization on the separation ground plate to raise the fundamental frequency of the stowed assembly from $34\,\text{Hz}$ to $84\,\text{Hz}$, above the $70\,\text{Hz}$ requirement of the lander, within the $1.6\,\text{kg}$ mass allocation.
\end{enumerate}

\section{Literature Review}

This section reviews the existing literature on planetary rover missions, payload integration, spacecraft-rover mechanical separation systems, and their experimental validation methodologies.

Planetary exploration increasingly utilizes robotic rovers to mitigate parent spacecraft landing risks and perform detailed surface science. For JAXA's (Japan Aerospace Exploration Agency) Martian Moons Exploration (MMX) mission, CNES (the French space agency) and DLR (the German Aerospace Center) are developing the joint IDEFIX rover, of approximately $25\,\text{kg}$ ($29\,\text{kg}$ including its support and separation system), to explore Phobos. Bertrand et al.~\cite{bertrand2019roving} detailed its ballistic deployment strategy, where the vehicle is passively released from the parent spacecraft at an altitude of about $50\,\text{m}$~\cite{delmaskrause2025mmx} using a bottom-mounted hold-down and release mechanism (HDRM) interfacing with the spacecraft's science panel. Because Phobos's very low gravity makes traditional rocker-bogie suspensions susceptible to slipping or rolling, IDEFIX employs a simple box chassis with four deployable rigid legs, non-directional wheels, and folded solar arrays to safely transport its scientific payload including the RAX Raman spectrometer and the miniRad radiometer~\cite{delmaskrause2025mmx}.

On larger scales, India's Chandrayaan-2 was designed to deploy the Pragyan rover from the Vikram lander via a post-landing ramp~\cite{sundararajan2018overview}, although the lander was lost during descent before the rover could be deployed, whereas the heavy Mars Science Laboratory (MSL) relied on an active ``Sky Crane'' tethered descent to deliver Curiosity under strict clearance constraints~\cite{prakash2008mars}. To optimize multi-mission utility, modular frameworks like the European Moon Rover System (EMRS)~\cite{luna2023european} adopt standardized locomotion buses flanked by interchangeable lateral and top payload modules. For micro-rovers, mass and volume envelopes are highly restricted. The $18\,\text{kg}$ polar MoonRanger micro-rover~\cite{schweitzer2022micro} maps volatiles using a neutron spectrometer, employing a suspension-less, 4-wheel skid-steer chassis to navigate autonomously outside lander communications. China's Chang'e-6 mission~\cite{yang2025technical} successfully deployed an ultra-lightweight, autonomous intelligent microrobot from the lander's side for cooperative imaging and farside surface exploration. Similarly, Tohoku University and ispace developed Koguma, a $1\,\text{kg}$ three-wheeled lunar micro-rover prototype~\cite{laine2018initial}, and evaluated its wheel and grouser geometries in slope-climbing tests on sand under tight mass and torque limits. At the other end of the small-rover range, payloads such as the Lunar Excavation and Size Separation System (LES3) on the $40\text{--}60\,\text{kg}$ LUVMI-X rover illustrate how tightly the traction and force budgets of lightweight platforms constrain their mechanisms, with an $8\,\text{N}$ horizontal force threshold imposed by low traction limits~\cite{just2022development}.

Micro-rovers that may land in an arbitrary attitude also benefit from mobility concepts that support self-righting. Siegwart et al.~\cite{siegwart1998design} proposed a ``stepping triple wheels'' micro-rover, combining rolling with a stepping sequence via twin tri-wheel frames that rotate independently around the chassis, allowing the rover to lift wheels over obstacles and shift its center of gravity (CoG) to maintain balance and recover its attitude. For highly loose soils, Patel~\cite{patel2005optimal} analyzed the Elastic Loop Mobility System (ELMS), whose continuous flexible bands provide a large contact area and low ground pressure, and developed design tools to optimize such mobility systems. Robotic surface operations also involve their own mechanical separation and transfer interfaces: Chang'e-5~\cite{xiao2021change} mechanically separated, packaged, and transferred drilled and scooped regolith samples from the lander to the ascent vehicle.

Vehicle-host mechanical separation is highly sensitive to tipoff disturbances. Palmer and Mitchell~\cite{palmer1966analysis} established mathematical models for multi-spring separation, showing that precise matching of helical compression springs is vital to prevent lateral forces and minimize angular tipoff errors. For on-orbit launches, Ma et al.~\cite{ma2026design} demonstrated that unilateral impact reaction forces induce severe torque disturbances along the roll axis. They showed that a cylindrical mechanism with continuous-contact straight guide rails ($270\,\text{mm}$ length, $0.2\,\text{mm}$ fit clearance) suppresses these disturbances by $82.76\%$ compared to gradient rails. Incorporating these dynamics, Jin et al.~\cite{jin2026rapid} simulated 1,200 spring preload and thrust misalignment combinations via Monte Carlo methods, validating that straight-rail assemblies restrict attitude deviation and suppress tipoff angular velocities.

To protect sensitive spacecraft electronics from pyrotechnic shocks, non-pyrotechnic low-shock alternatives are increasingly preferred. Dowen et al.~\cite{dowen2001development} developed a reusable clamp band system using Fast Acting Shockless Separation Nut (FASSN) technology to convert stored strain energy into flywheel rotational energy, reducing peak shocks to under $400\,\text{g}$. Lee and Han~\cite{lee2024separation} reviewed release actuator taxonomies, contrasting pyrotechnics with smart-material actuators (SMAs, paraffin, split-spools) across load, timing, and resettability. Smith et al.~\cite{smith1999development} demonstrated SMA-based Qwknut and Low-Force Nuts (LFN) for non-explosive releases up to 3,000~lbf, while Fouch\'{e} et al.~\cite{fouche2017development} verified high-temperature CN-250X SMA actuators triggering above 120$^\circ$C for shock-free release. These low-shock principles extend to surface-bound rovers. Historically, the Apollo Lunar Roving Vehicle (LRV) deployment system~\cite{hunter1972lunar} utilized semi-automatic Space Support Equipment with $\text{MoS}_2$-lubricated worm gears and a braked-reel linkage to passively unfold and lower the 210~kg vehicle within a $120\,\text{N}$ astronaut force limit. Modern micro-rovers adapt these passive concepts; MoonRanger~\cite{schweitzer2022micro} employs a gravity-assisted ``drop deployment'' where the vehicle freefalls 1~meter from bottom-mounted zero-force HDRMs. Conversely, Chang'e-3~\cite{liu2014rover} used an active folding ladder transfer system with spring-loaded hinges and a cable-driven winch to lower Yutu. This mechanism deployed the rover about seven hours after landing, with lander and rover cameras capturing mutual images to confirm separation and structural integrity~\cite{ye2014change}.

Dynamic verification campaigns often rely on reusable substitutes and specialized mechanical ground support equipment (MGSE). For the MMX separation campaign, Springer~\cite{springer2022design} replaced flight-grade melted-wire HDRMs with mechanical wrench-triggered versions actuated by drop-weights and pulleys to enable cost-effective testing. To model these dynamics, Pign\`{e}de et al.~\cite{pignede2025mmx} performed 100,000-sample Monte Carlo sensitivity analyses on the IDEFIX chassis separation, revealing that post-ejection tumbling rates are highly sensitive to mass offsets and establishing center-of-mass alignment as a critical separation constraint. To replicate low gravity, both the MMX tests~\cite{springer2022design} and the classical Vela/OGO campaigns~\cite{palmer1966analysis} utilized long rope pendulums to suspend the chassis, providing three rotational degrees of freedom while minimizing vertical error. However, because passive suspensions introduce cable drag and friction, active gravity offloaders---such as NASA's Active Response Gravity Offload System (ARGOS)~\cite{nasa2015active} and Carnegie Mellon's SM2 gravity-compensation system~\cite{nechyba1995sm2}---use closed-loop force tracking to keep the offload force nearly constant during motion. To validate complex operations, analog campaigns like LUNARES~\cite{cordes2011lunares} utilize simulated craters with 30$^\circ$--45$^\circ$ slopes to verify visual-servoing docking and sample transfer between heterogeneous rovers.

Kinematics of passive deployments require targeted environmental and ground testing. The MoonRanger mission~\cite{schweitzer2022micro} validated its gravity-drop deployment via freefall testing and dynamic terramechanics simulations to ensure upright landing. Furthermore, the abrasive lunar environment demands rigorous environmental qualification; Liu et al.~\cite{liu2014rover} subjected Yutu's transfer hinges to thermal-vacuum (TVAC) cycling and basalt-dust blowing tests ($20\,\mu\text{m}$ to $1\,\text{mm}$), proving that rotating seals prevent dust ingress and friction rise. Returned farside regolith from Chang'e-6~\cite{qi2025strongly} exhibits exceptionally high cohesion (static angle of repose up to $52.9^\circ$, dynamic up to $70.4^\circ$) due to plagioclase abundance and space weathering, highlighting the importance of representative regolith simulants when testing the touchdown and settling of deployed micro-rovers.

In summary, the state-of-the-art demonstrates that spacecraft-rover separation and deployment systems are inherently bespoke, with each architecture highly optimized for its specific parent spacecraft interface, target body gravity, and rover scale. While heavy-class missions rely on high-energy active or complex motorized transfer mechanisms, micro-rovers demand minimalist, low-mass, and low-shock alternatives. However, little published work presents a comprehensive design methodology that covers the entire development lifecycle of micro-rover hold-down and separation systems---spanning from initial kinematic and spring-force optimization to multi-body dynamic simulations under parametric uncertainties and cost-effective ground testing campaigns. To address this gap, this paper details the design, analysis, and experimental validation of a lightweight, low-shock separation system for a lunar micro-rover. By combining multi-body dynamic simulation with weight-matched ground tests, and by stating explicitly which aspects of flight behavior such tests can and cannot reproduce, this work provides a transferable framework for the development and testing of future micro-rover deployment mechanisms.

\section{Mechanical Architecture and Design Evolution}
\label{sec:mechanical_design}

To transition the CHERI micro-rover from a stowed launch configuration to nominal surface locomotion, the mechanical separation assembly must execute a reliable, low-shock ejection. This section details the system's physical design, the engineering constraints that drove its development, and the iterative design journey that culminated in the final tested mechanical architecture.

\subsection{Design Requirements and Interface Constraints}
The mechanical separation system is governed by stringent mass, volume, and stiffness constraints imposed by the lunar lander's payload bay and the launch vehicle's dynamic environment. The key engineering requirements are formalized as follows:
\begin{enumerate}
    \item \textbf{Mass Allocation:} The total mass of the separation assembly---including the ground plate, hold-down and release mechanism (HDRM), springs, and guide housings---must not exceed $1.2\,\text{kg}$, representing $24\%$ of the $5.0\,\text{kg}$ rover mass. This budget was relaxed to $1.6\,\text{kg}$ ($32\%$ of the rover mass) during the separation-nut era (Iterations 4--6) to accommodate the heavier separation nut, the guide-housing tribological reinforcements, and the stiffening of the ground plate (Table~\ref{tab:design_evolution_comparison}).
    \item \textbf{Kinematic Clearance:} The system must propel the rover horizontally by at least $D_{\text{clearance}} \ge 0.7\,\text{m}$ upon ejection from a lateral stowed height of $H = 1.7\,\text{m}$, ensuring complete physical clearance of the lander's primary structural footpads.
    \item \textbf{Attitude Stability in Flight:} The separation must not make the rover tumble: its pitch and roll angles must remain within $\pm 5^\circ$ during ejection and free flight, since larger rotations can lead to an inverted landing. This is a kinematic performance target for the separation mechanism itself, governed by the spring array's residual pitch rate, rather than a bound on the settled state after bounce and roll on sloped terrain, which is characterized separately in Section~\ref{sec:experimental_validation}.
    \item \textbf{Structural Resonance:} To survive launch-induced sinusoidal and random vibrations without structural coupling, the separation plate assembly must exhibit a first natural frequency of $f_1 \ge 70\,\text{Hz}$ when preloaded with the stowed rover mass, as required by the lander specification for the launch and landing phases~\cite{bice2025devspec}, so that its modes remain decoupled from the low-frequency launch excitation~\cite{sarafin1995spacecraft}.
\end{enumerate}

\begin{figure}[htbp]
\centering
\includegraphics[width=0.48\textwidth]{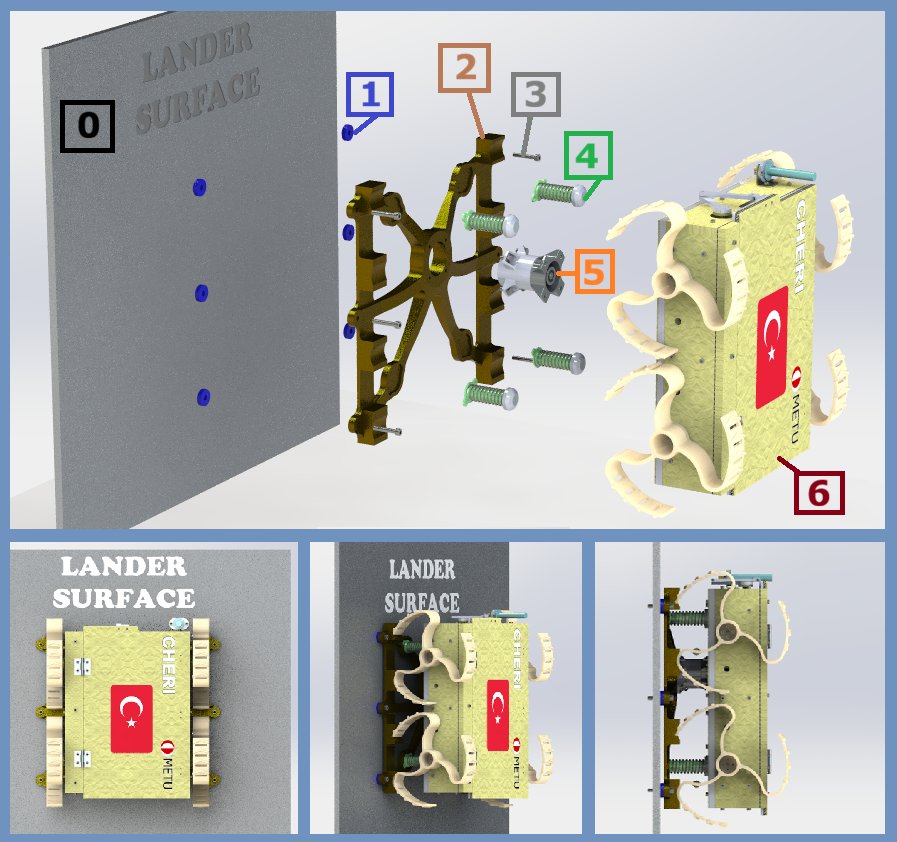}
\caption{Labeled exploded assembly view of the CHERI micro-rover separation interface, detailing the key physical components and sequence of structural integration: (0) lander surface, (1) isolation pads, (2) ground plate, (3) mounting screws, (4) separation springs, (5) separation nut, and (6) CHERI Rover.}
\label{fig:exploded_assembly}
\end{figure}

The detailed mechanical design and assembly stack-up of the separation system, showing the precise layers from the lander surface up to the stowed vehicle, is illustrated in the labeled exploded view in Fig.~\ref{fig:exploded_assembly}.

\subsection{Evolutionary Iterations of the Ground Plate and Actuator Layout}
The final separation system was not achieved in a single design cycle; rather, it was the product of six distinct evolutionary design generations, each addressing failures identified during laboratory and shaker-table testing. Every iteration described below was manufactured and tested.

\begin{figure*}[htbp]
\centering
\setlength{\tabcolsep}{2pt}
\begin{tabular}{ccc}
\includegraphics[width=0.25\textwidth]{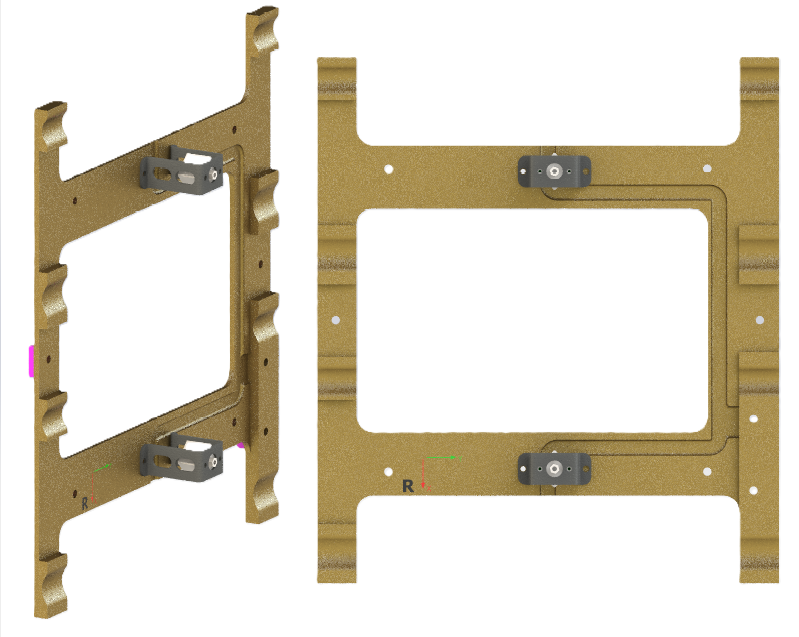} &
\includegraphics[width=0.32\textwidth]{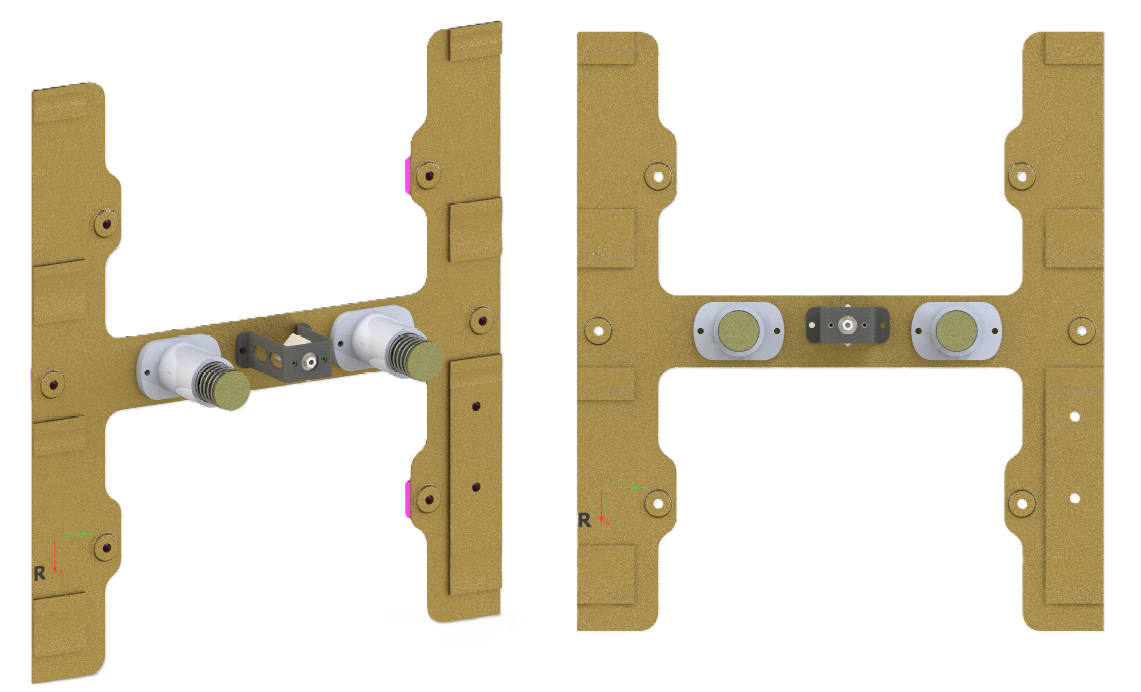} &
\includegraphics[width=0.32\textwidth]{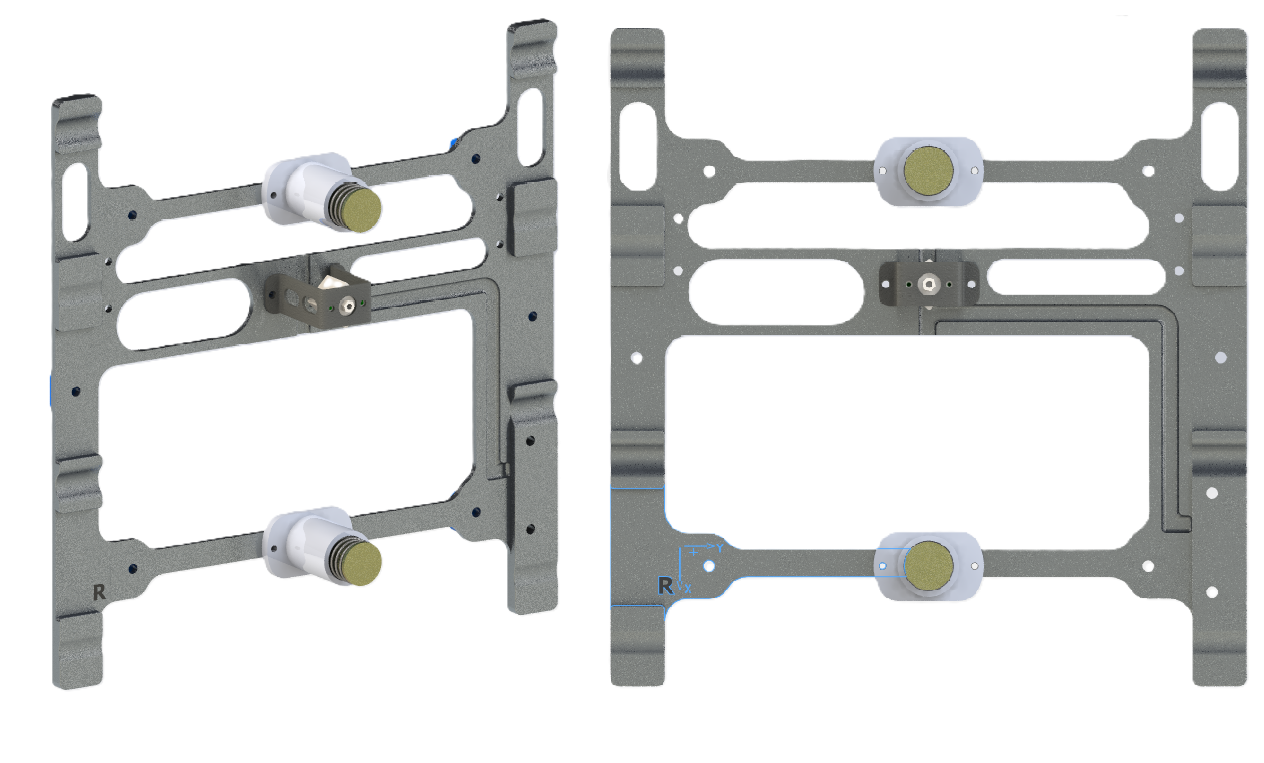} \\
(a) Iteration 1 CAD Model & (b) Iteration 2 CAD Model & (c) Iteration 3 CAD Model \\ [6pt]
\includegraphics[width=0.26\textwidth]{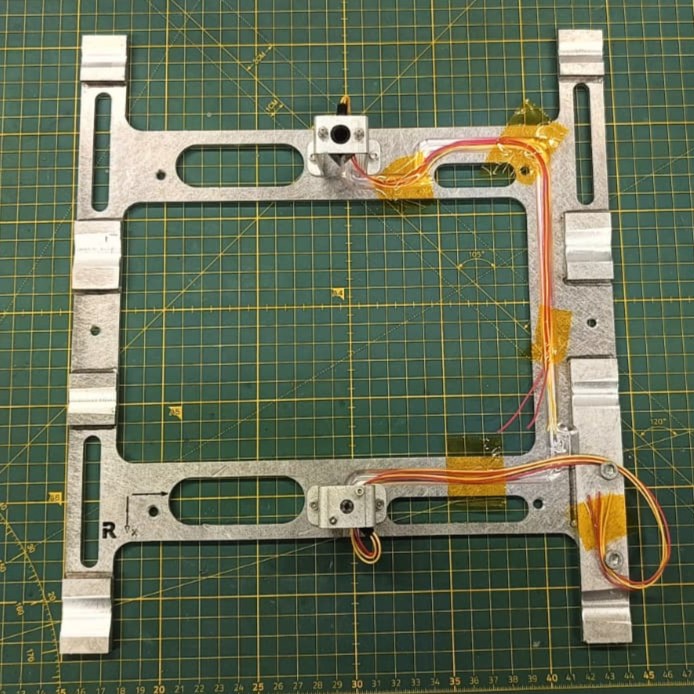} &
\includegraphics[width=0.30\textwidth]{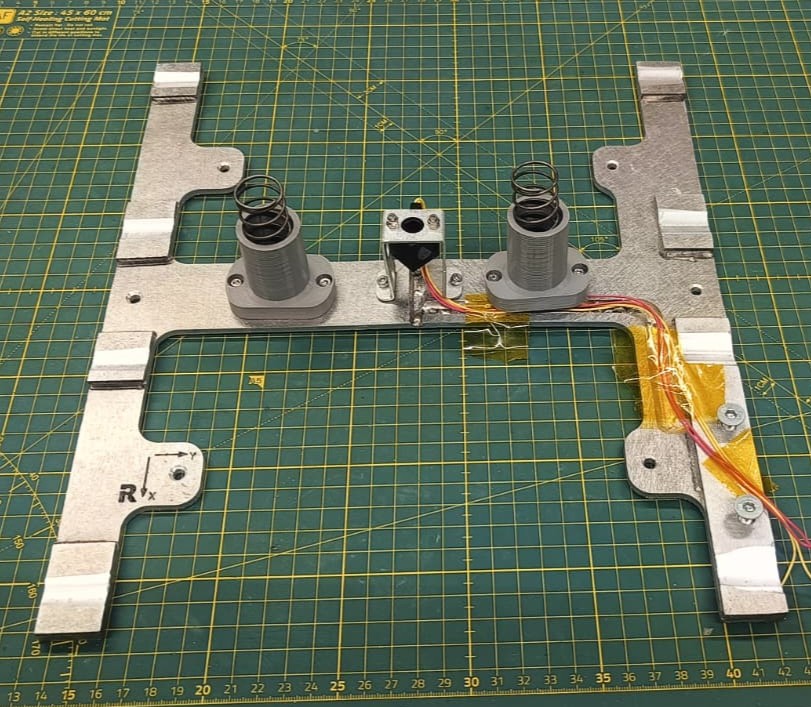} &
\includegraphics[width=0.32\textwidth]{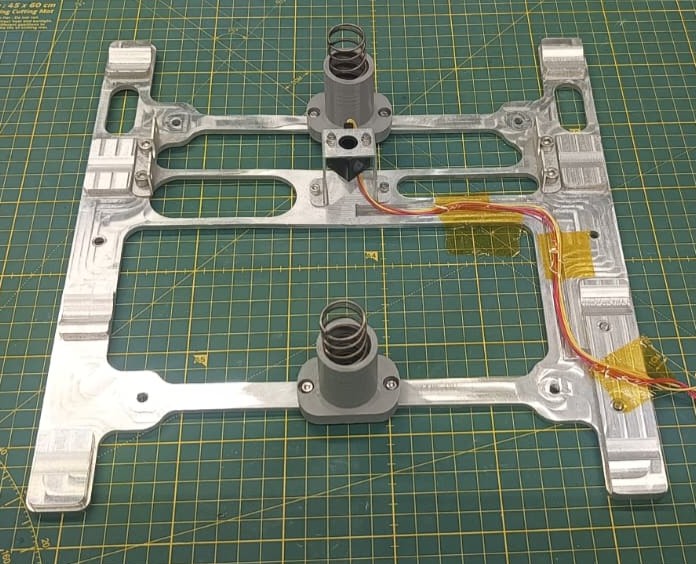} \\
(d) Iteration 1 Physical Setup & (e) Iteration 2 Physical Setup & (f) Iteration 3 Physical Setup
\end{tabular}
\caption{First technological era using micro shape memory alloy (SMA) release nuts (Iterations 1--3): (a--c) isometric CAD models; (d--f) manufactured hardware prototypes.}
\label{fig:release_nut_era}
\end{figure*}

The mechanical development is chronologically divided into two major technological eras:
\begin{enumerate}
    \item \textbf{The Micro Shape Memory Alloy (SMA) Release Nut Era (Iterations 1--3):} Characterized by ultra-lightweight designs optimized for low preloads, where vertical and horizontal alignment variations were evaluated using a micro shape memory alloy release nut~\cite{nD3RN_datasheet} (shown in Fig.~\ref{fig:release_nut_era}).
    \item \textbf{The Fracture-Type Separation Nut Era (Iterations 4--6):} Marked by a transition to heavy-duty preloads and specialized guide/structural optimizations to resolve stiction and resonance challenges (shown in Fig.~\ref{fig:separation_nut_era}).
\end{enumerate}

\subsubsection{First Generation (Iteration 1): Dual-Actuator H-Plate}
The first-generation design was physically manufactured as a prototype utilizing a lightweight ground plate with an H-profile consisting of two horizontal support bars, as illustrated in Fig.~\ref{fig:release_nut_era}(a) and manufactured in Fig.~\ref{fig:release_nut_era}(d). The deployment strategy relied on two active, synchronized micro SMA release nuts positioned symmetrically relative to the vehicle's horizontal axis. To achieve a flat touchdown attitude, the design proposed firing the two actuators with a millisecond-level delay, so that gravity would generate a corrective pitching moment during release.

However, prototype evaluations revealed two critical flaws:
\begin{itemize}
    \item \textbf{Synchronization Risk:} The timed firing of two release nuts was highly sensitive to electrical and mechanical jitter. A delay mismatch, or the failure of either actuator, would generate an uncontrolled tumbling moment; each actuator was therefore a single point of failure.
    \item \textbf{No Ejection Capability:} Release nuts are release-only devices and provide no pushing force. Without springs, the rover would leave the plate under gravity alone, with no horizontal velocity to reach the $0.7\,\text{m}$ clearance, while the duplicated actuators and electronics added mass and complexity.
\end{itemize}

\subsubsection{Second Generation (Iteration 2): Single Actuator with Dual Passive Springs}
To eliminate the synchronization risk, the second generation reduced the active component to a single, central micro SMA release nut aligned with the rover's Center of Mass (CoM). This active unit was flanked by two passive, helical compression springs on the left and right sides to provide symmetric ejection force. The ground plate was redesigned as a single-bar H-profile, shown in Fig.~\ref{fig:release_nut_era}(b) and pictured in Fig.~\ref{fig:release_nut_era}(e), and manufactured using Al 7075-T6.

During experimental test firings, this configuration failed to meet the clearance requirements. With high-friction structural interfaces, the pushing force of the two flanking springs was insufficient to propel the rover beyond the $0.7\,\text{m}$ lander clearance envelope, and the vehicle clipped the lander's simulated leg.

\subsubsection{Third Generation (Iteration 3): Three-Bar Plate with Micro Release Nut}
To increase the ejection energy, the ground plate was evolved into a three-horizontal-bar profile, as shown in Fig.~\ref{fig:release_nut_era}(c) and manufactured in Fig.~\ref{fig:release_nut_era}(f). This design moved the two springs from the side-by-side arrangement of Iteration 2 to positions above and below the central release nut, with their axes still normal to the plate. To support the increased spring stiffness while maintaining the lightweight design philosophy, the central micro SMA release nut was retained for this third iteration.

While this layout successfully increased the potential energy storage, experimental characterization under cyclic test conditions revealed that this configuration was unsuitable for the repetitive testing required for this multi-spring setup:
\begin{itemize}
    \item \textbf{Inadequacy under Cyclic Test Conditions:} Although the micro SMA release nut is specified as a reusable device, preliminary trials showed that, under the installation torques and bending moments imposed by the stiffer spring array, it lacked the cycle-to-cycle repeatability required for the many ground-test runs of this multi-spring system (Section~\ref{sec:analytical_optimization}). This prompted the transition to a more robust separation nut.
\end{itemize}

\begin{figure*}[htbp]
\centering
\setlength{\tabcolsep}{2pt}
\begin{tabular}{ccc}
\includegraphics[width=0.32\textwidth]{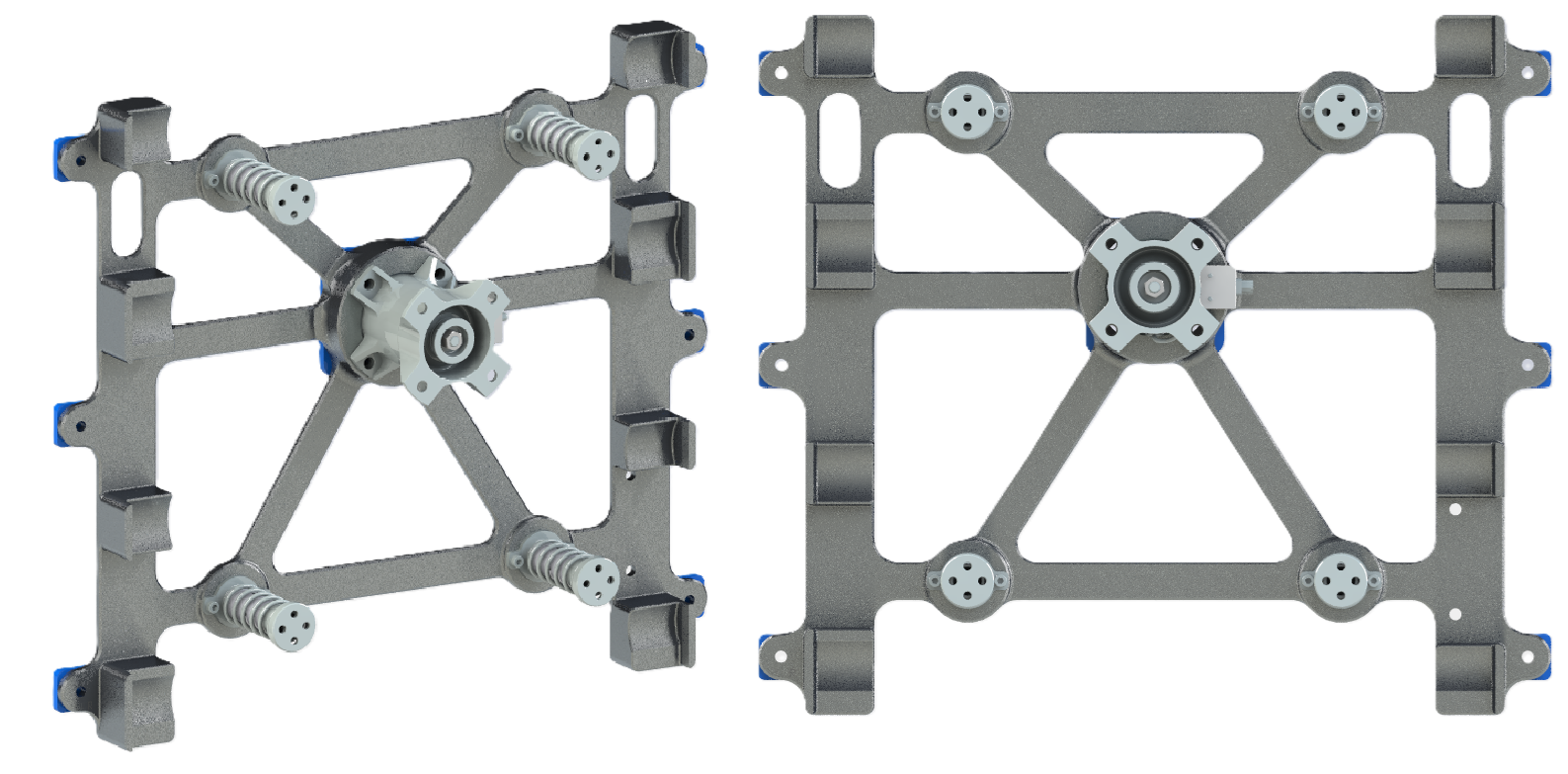} &
\includegraphics[width=0.32\textwidth]{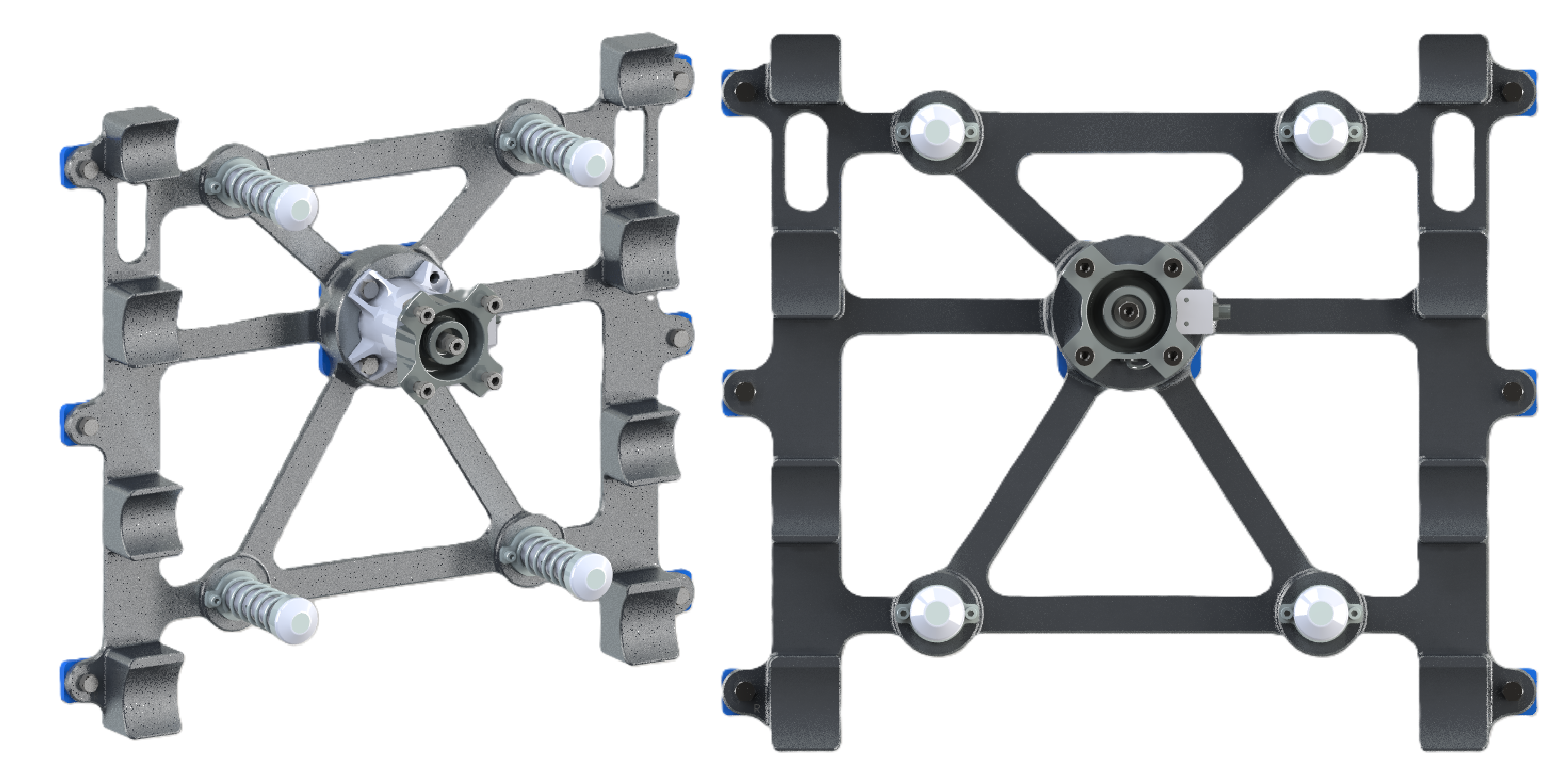} &
\includegraphics[width=0.32\textwidth]{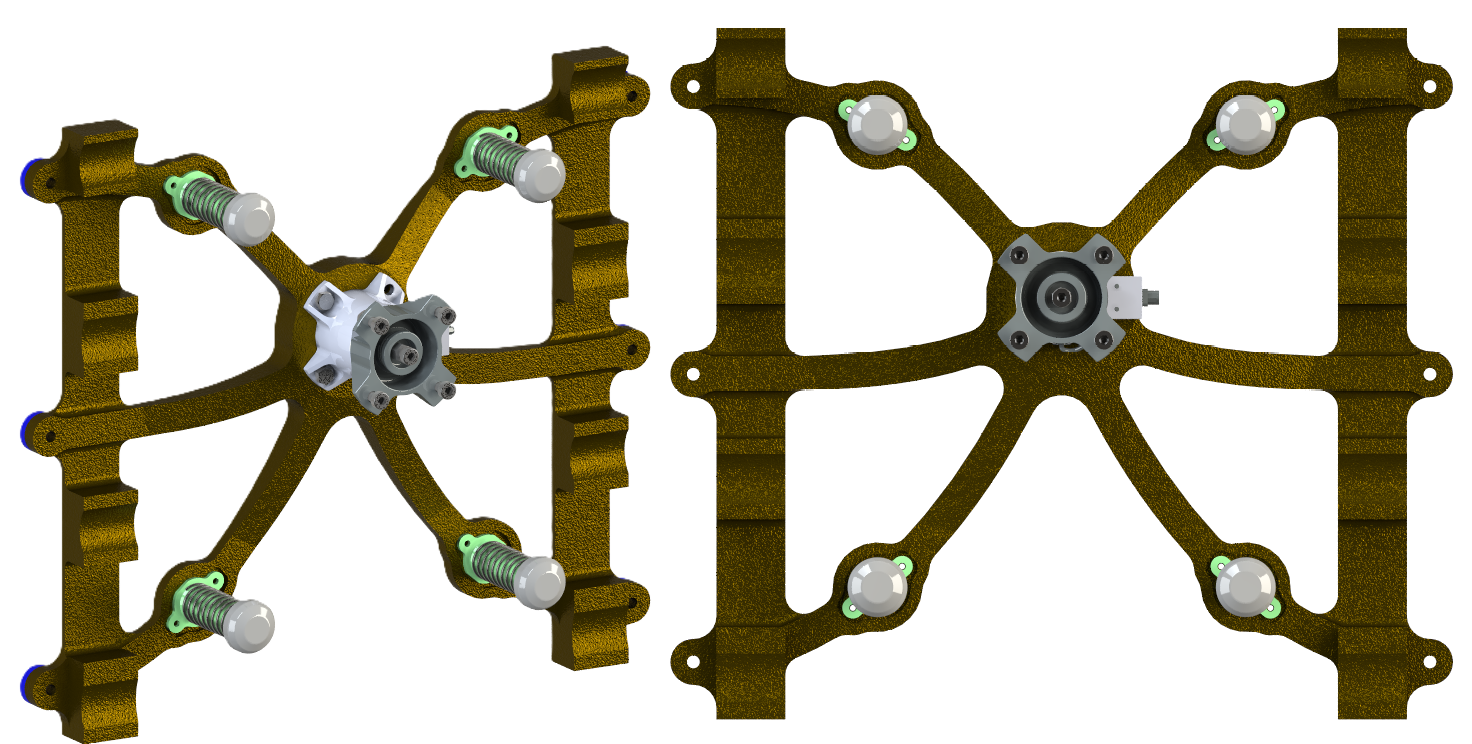} \\
(a) Iteration 4 CAD Model & (b) Iteration 5 CAD Model & (c) Iteration 6 CAD Model \\ [6pt]
\includegraphics[width=0.27\textwidth]{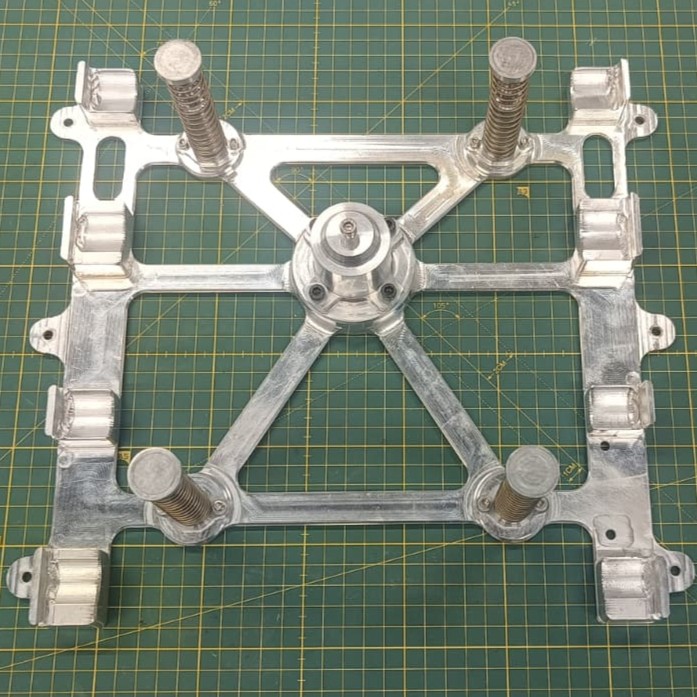} &
\includegraphics[width=0.30\textwidth]{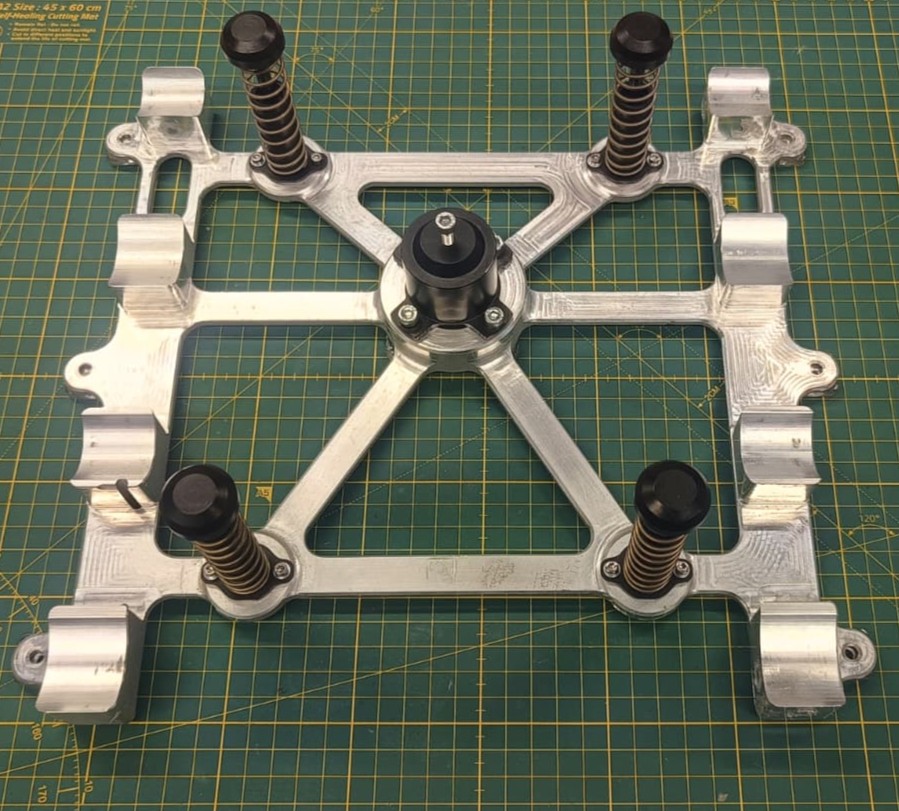} &
\includegraphics[width=0.33\textwidth]{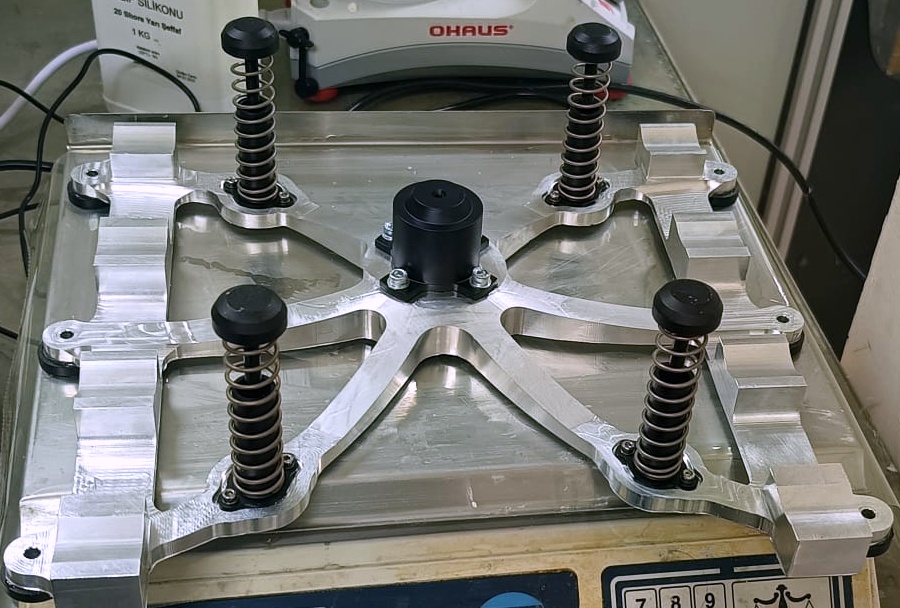} \\
(d) Iteration 4 Physical Setup & (e) Iteration 5 Physical Setup & (f) Iteration 6 Physical Setup
\end{tabular}
\caption{Second technological era using a fracture-type separation nut~\cite{sfr_datasheet} (Iterations 4--6): (a--c) isometric CAD models; (d--f) corresponding physical hardware.}
\label{fig:separation_nut_era}
\end{figure*}

\subsubsection{Fourth Generation (Iteration 4): Pre-Optimization 4-Spring Layout}
To overcome the payload preloading limits, the fourth-generation design abandoned the micro release nut and transitioned to a robust, non-explosive separation nut, capable of withstanding preloads up to $5\,\text{kN}$, as depicted in Fig.~\ref{fig:separation_nut_era}(a) and manufactured in Fig.~\ref{fig:separation_nut_era}(d). The spring array was completely redesigned to feature four heavy-duty helical compression springs.

The spring layout is symmetric about the horizontal (X) axis to avoid parasitic yaw rates, but asymmetric along the vertical (Y) axis. Offsetting the resultant spring force from the rover's CoM imparts a controlled pitching rate during separation, intended to counteract the tip-off rotation that gravity induces while the laterally stowed rover is still in contact with the plate, and to bias the touchdown attitude toward wheels-first.

During separation trials, the horizontal separation velocity fell short of the analytical energy-conservation estimate. Video analysis indicated lateral buckling and sliding stiction of the springs against the unoptimized guide rods, which absorbed part of the stored energy as friction. In the first vibration campaign at T\"{U}B\.{I}TAK UZAY, the unbalanced spring energies and guide tolerances caused one spring to jam in its guide housing during Z-axis random vibration, producing excessive dynamic displacement that halted the test (Section~\ref{sec:analytical_optimization}). This first campaign was a development test: the first natural frequency of $51\,\text{Hz}$ measured in the low-level Z-axis sine sweep was recorded but not used as a stop criterion.

\subsubsection{Fifth Generation (Iteration 5): Tribologically-Optimized Spring Guides}
To resolve the friction losses and jamming without changing the plate design, Iteration 5 used the same plate geometry as Iteration 4 with a re-engineered spring guide housing and rod assembly and the re-optimized $85.0\,\text{mm}$ springs of Section~\ref{sec:analytical_optimization}, as shown in Fig.~\ref{fig:separation_nut_era}(b) and pictured as an assembled unit in Fig.~\ref{fig:separation_nut_era}(e).

The optimization of the guide mechanism integrated:
\begin{itemize}
    \item \textbf{Material Pairing:} Dissimilar metal pairing, using precision-ground stainless steel rods (AISI 304) mating with precision-machined aluminum bushings (Al 7075-T6) to prevent cold-welding.
    \item \textbf{Hard Anodization:} Hard anodization of the sliding surfaces of the aluminum bushings to maximize wear resistance.
    \item \textbf{Dry-Film Lubrication:} A Molykote molybdenum disulfide ($\text{MoS}_2$) dry-film coating with a manufacturer-specified sliding friction coefficient of $\mu \le 0.08$. With this treatment, no stiction or jamming was observed in the subsequent separation trials.
\end{itemize}

In the second vibration campaign, the spring guides performed nominally, but the low-level sine sweep measured the first natural frequency of the stowed assembly at $34\,\text{Hz}$, far below the $70\,\text{Hz}$ requirement, and the test was stopped. The failure was traced to flexing of the flat ground plate (Section~\ref{sec:structural_dynamics}).

\subsubsection{Sixth Generation (Iteration 6): Final Topology-Optimized Plate}
To resolve the stiffness-to-weight deficit of the fifth-generation prototype, the ground plate was completely re-engineered in Al 7075-T6 using Solid Isotropic Material with Penalization (SIMP) topology optimization, as shown in Fig.~\ref{fig:separation_nut_era}(c) and shown as the final tested hardware in Fig.~\ref{fig:separation_nut_era}(f). The optimization redistributed material to increase bending and torsional stiffness while pocketing out low-strain regions.

When manufactured and subjected to dynamic validation campaigns at the
satellite testing facilities, this sixth-generation, topology-optimized
ground plate raised the assembly's first natural frequency to
$84\,\text{Hz}$, above the $70\,\text{Hz}$ threshold. The updated
hardware then passed sinusoidal and random vibration and pneumatic
shock testing at the specified levels without structural degradation. The complete evolutionary history of the six generations
spanning actuator type, spring configuration, mass, natural frequency, and
critical failure mode is summarized in
Table~\ref{tab:design_evolution_comparison}.

\begin{table*}[htbp]
\centering
\caption{System-level engineering comparison of the six evolutionary separation system generations.}
\label{tab:design_evolution_comparison}
\footnotesize
\setlength{\tabcolsep}{4pt}
\begin{tabular}{clccllc}
\toprule
\textbf{Iter.} & \textbf{Actuator Type} & \textbf{Spring Config.} & \textbf{Mass (kg)} & \textbf{1st Nat. Freq.} & \textbf{Critical Failure Mode} & \textbf{Status} \\
\midrule
1 & 2 $\times$ Release Nut & None (Gravity) & 0.85 & NA & Actuator synchronization; no ejection force & NA \\
2 & 1 $\times$ Release Nut & 2 $\times$ Lateral & 0.92 & NA & Insufficient clearance (clips lander footpad) & NA \\
3 & 1 $\times$ Release Nut & 2 $\times$ Vertical & 1.05 & NA & Inadequacy under cyclic test conditions & NA \\
4 & 1 $\times$ Separation Nut & 4 $\times$ Asymmetric & 1.28 & 51\,\text{Hz} & Spring stiction; spring jam in Z-axis random vib. & Failed \\
5 & 1 $\times$ Separation Nut & 4 $\times$ Asym. (Opt. Guides) & 1.48 & 34\,\text{Hz} & First mode below lander requirement (70\,\text{Hz}) & Failed \\
\textbf{6} & \textbf{1 $\times$ Separation Nut} & \textbf{4 $\times$ Asym. (Opt. Guides)} & \textbf{1.54} & \textbf{84\,\text{Hz}} & \textbf{None (SIMP topology-optimized plate)} & \textbf{Passed} \\
\bottomrule
\end{tabular}
\end{table*}

\subsection{Lander Interface Design System (IDS) and Thermal Isolation}
To couple the ground plate to the lander, a dedicated Lander Interface Design System (IDS) is established. The ground plate is physically mounted to the lander structural frame using a total of six M5 bolts, arranged symmetrically with three bolts on the left side and three on the right side of the plate. To mitigate conductive heat transfer between the lunar lander's primary structure and the separation assembly, a circular polyimide isolation spacer with an outer diameter of $\varnothing 22\,\text{mm}$ and a thickness of $5\,\text{mm}$ is placed at each of the six bolt locations, directly sandwiched between the ground plate and the lander mating surface. Polyimide was selected for its low thermal conductivity, its stability over a wide temperature range (from cryogenic temperatures to well above lunar-day temperatures), and its high compressive strength under launch preloads.

These high-performance polymeric spacers serve a dual engineering purpose:
\begin{enumerate}
    \item \textbf{Conductive Decoupling:} They reduce the conductive contact area between the metallic ground plate and the lander body, thermally isolating the stowed rover from the temperature fluctuations of the lander structure.
    \item \textbf{Load Distribution:} The $5\,\text{mm}$ pads carry the bolt preloads and distribute the contact load at the bolt interfaces, reducing local stress concentrations in the plate during launch.
\end{enumerate}

The reference origin of the Lander Interface Design System (IDS) coordinate system, designated as the R-point (Reference Point), is defined as the center of the bottom-left mounting hole of the ground plate. This R-point serves as the primary alignment datum for the physical interface layout with the lander. Crucially, while the IDS layout and lander attachment coordinates are referenced to the R-point, the separation kinematics, spring forces, and angular moment balances of the ejection system (detailed in Section~\ref{sec:analytical_optimization}) are calculated relative to the central separation nut, whose axis passes through the rover's nominal Center of Mass (CoM) so that the hold-down itself introduces no tumbling torque.

\subsection{Self-Righting Recovery via Leg Maneuvers and the Deployable Solar Panel}
While the optimized 4-spring asymmetric separation system is designed to promote a flat, wheels-first touchdown, terrain irregularities, boulders, steep slopes, or the residual pitch rate of the separation event itself can still cause the vehicle to land in an inverted state; the conditions under which this occurs are characterized in Section~\ref{sec:experimental_validation}. To preserve mission capability in that case, the CHERI micro-rover incorporates a secondary, active self-righting recovery system.

Because of strict micro-rover mass constraints, this recovery system avoids dedicated flipping hardware and uses existing subsystems instead:
\begin{itemize}
    \item \textbf{Coordinated Leg Maneuvers:} The rover's four J-shaped flexible wheel-legs are driven by independent motors. If an inverted state is detected by the onboard Inertial Measurement Unit (IMU), the rover executes a pre-programmed, coordinated leg swing, rotating its limbs to change its contact geometry with the regolith.
    \item \textbf{Single Deployable Solar Panel:} Simultaneously, the rover actuates the deployment hinge of its single, top-mounted solar panel, which acts as a lever pushing against the regolith.
\end{itemize}
By combining the vertical lift generated by the single unfolding panel with the momentum shift from the coordinated leg maneuvers, the rover shifts its Center of Gravity (CoG) outside its inverted support polygon and flips the chassis back onto its wheels. Quantifying this capability is ongoing work (Section~\ref{sec:experimental_validation}).

\section{Analytical Kinematics and Separation Optimization}
\label{sec:analytical_optimization}

To guarantee a reliable dynamic separation that clears the lander footpads while maintaining attitude stability, the force profiles and stiffness characteristics of the ejection springs must be modeled analytically. This section details the initial sizing using commercial components, the experimental discovery of parasitic energy absorption within the rover's flexible mobility limbs, the multi-body simulations in MuJoCo, and the subsequent numerical solver-based optimization of the spring parameters that resolved the dynamic failures.

\subsection{Initial Sizing and Commercial Off-The-Shelf (COTS) Baseline}
The initial design of the separation mechanism was scaled under rigid-body assumptions. The primary kinematic objective was to propel the $5.0\,\text{kg}$ micro-rover horizontally to clear the $D_{\text{clearance}} \ge 0.7\,\text{m}$ envelope from a mounting height of $H = 1.7\,\text{m}$. Under lunar gravity ($g_{\text{Moon}} = 1.62\,\text{m/s}^2$), the fall time is $\sqrt{2H/g_{\text{Moon}}} \approx 1.45\,\text{s}$, and the required initial horizontal separation velocity of the CoM is:
\begin{equation}
v_x \ge \frac{D_{\text{clearance}}}{\sqrt{2H/g_{\text{Moon}}}} = \frac{0.7}{\sqrt{2(1.7)/1.62}} \approx 0.48\,\text{m/s}
\label{eq:vx_req}
\end{equation}
For the $1/6$ mass model tested under terrestrial gravity (Section~\ref{sec:gravity_scaling}), the equivalent requirement is $v_x \ge 0.7/\sqrt{2(1.7)/9.81} \approx 1.19\,\text{m/s}$. Assuming a perfectly rigid chassis and neglecting gravitational effects during the short release phase, the energy balance is:
\begin{equation}
\frac{1}{2} K_{\text{sys}} \left(\delta_0^2 - \delta_r^2\right) = \frac{1}{2} m v_x^2
\end{equation}
where $K_{\text{sys}}$ is the equivalent stiffness of the spring array, $\delta_0$ and $\delta_r$ are the spring compressions in the stowed state and at the release (separation) point, and $m$ is the rover mass. Under these idealized assumptions, only $\frac{1}{2}(5.0)(0.48)^2 \approx 0.58\,\text{J}$ of delivered energy is required, and commercial springs of approximately $10\,\text{N}$ force each appeared sufficient.

To validate this baseline, initial free-fall tests were conducted utilizing commercial off-the-shelf (COTS) helical compression springs. During these tests, the stowed rover was held down by the central micro shape memory alloy (SMA) release nut~\cite{nD3RN_datasheet}. However, the COTS springs failed to propel the rover beyond the clearance boundary. Furthermore, the micro release nut~\cite{nD3RN_datasheet} proved conceptually and operationally unsuitable for this high-energy separation system. Because a release nut functions purely as a passive, release-only hold-down mechanism, it possesses no active pushing or ejection capability. 

To overcome the compliance of the J-legs and store sufficient ejection energy, considerably stiffer custom springs were required. Compressing this high-stiffness spring array subjected the central hold-down interface to significant preloading forces, high installation torques, and multi-directional bending moments. While the micro release nut operates nominally under pure axial tension within its design limits, its miniature, lightweight internal latching interfaces are highly sensitive to such coupled installation torques and bending disturbances. During repetitive ground trials, this sensitivity made it operationally inconsistent. Therefore, the design transitioned to a more robust, fracture-type separation nut~\cite{sfr_datasheet}. Since both the release nut and the separation nut are passive hold-down and release mechanisms lacking active pushing plungers, the ejection force is entirely supplied by the optimized custom four-spring array. The separation nut was selected because its heavier construction tolerates the installation torques and bending moments imposed by the compressed spring assembly; as a fracture-type device it is consumed at each actuation, so each test run uses a new unit and run-to-run consistency does not depend on reset kinematics.

\subsection{Custom Spring Parametrization and Damping Losses}
To address the inadequacy of the COTS components, an analytical numerical model was developed to design a custom helical compression spring array. The spring stiffness $k$ and solid length $L_s$ are defined by the standard elastic formulations:
\begin{equation}
k = \frac{G d^4}{8 D^3 n_a}
\end{equation}
\begin{equation}
L_s = n_t\, d
\end{equation}
where $G$ is the shear modulus of the spring material (stainless steel AISI 304, $G \approx 69\,\text{GPa}$), $d$ is the wire diameter, $D$ is the mean coil diameter (defined as $D = D_o - d$, with $D_o$ as the outer diameter), $n_a$ is the number of active coils, and $n_t$ is the total number of coils (squared and ground ends).

\begin{figure}[htbp]
\centering
\includegraphics[width=\columnwidth]{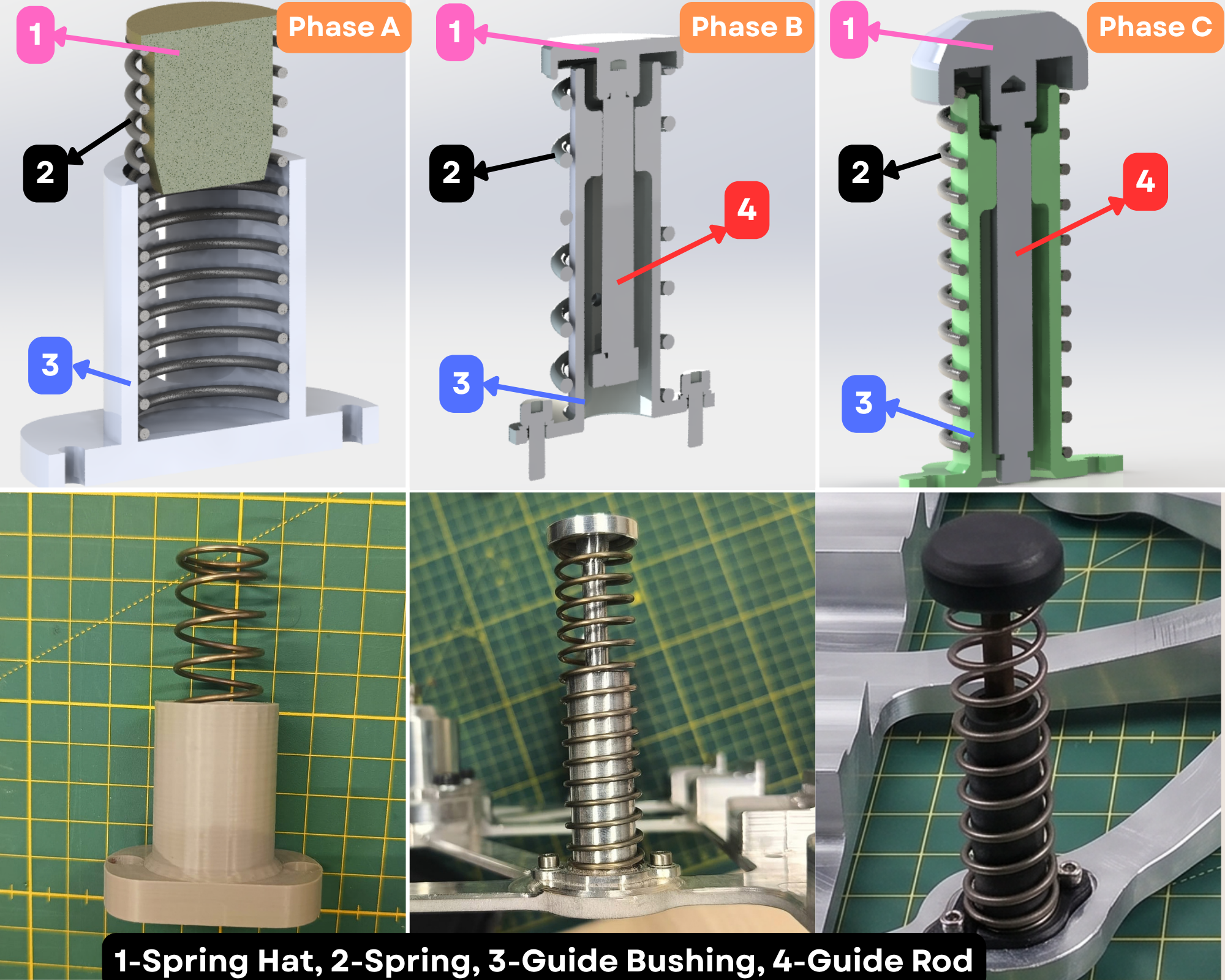}
\caption{Chronological evolution of the separation spring mechanisms, detailing CAD cross-sections (top row) and corresponding physical hardware (bottom row). Left (Phase A): the initial enclosed COTS-based baseline. Middle (Phase B): the first custom iteration, whose unoptimized guide allowed lateral buckling and sliding stiction during release. Right (Phase C): the final configuration with the re-engineered guide bushing and rod. Key components are annotated: (1) spring hat, (2) helical spring, (3) guide bushing, and (4) central guide rod. The re-engineered guide bushing (3) and rod (4) ensure a stable, linear stroke without buckling or stiction.}
\label{fig:spring_evolution_collage}
\end{figure}

Using this model, a custom spring set was designed by calculating the coil diameter, wire diameter, outer diameter, free length ($L_f$), and solid length ($L_s$). The manufactured custom springs were integrated into the assembly. However, initial drop testing revealed that the system behaved as if the springs were excessively ``soft,'' yielding almost no improvement in horizontal clearance and exhibiting lateral buckling under compression. The physical prototyping stages of these custom springs---from the inadequate initial baseline to the final guided and optimized set are shown in Fig.~\ref{fig:spring_evolution_collage}.

A diagnostic investigation showed that the rigid-body assumption was invalid. Under the compressive preloads of the stowed state, the rover's flexible J-shaped wheel-legs and chassis acted as parallel compliant and damping elements. During the release transient, a substantial portion of the stored elastic energy was absorbed by structural damping and by the elastic deformation of the J-legs:
\begin{equation}
E_{\text{loss}} = \int_{0}^{\delta_L} F_{\text{damp}}(\dot{x})\,dx + \frac{1}{2} k_{\text{leg}} x_{\text{leg}}^2
\end{equation}
where $F_{\text{damp}}(\dot{x})$ is the equivalent damping force of the chassis and legs, $\delta_L$ is the spring stroke up to the release point, and $k_{\text{leg}}$ and $x_{\text{leg}}$ are the equivalent stiffness and deflection of the J-legs. Instead of accelerating the vehicle's Center of Mass (CoM), part of the spring energy was spent deforming the flexible wheel-legs, which acted as an energy sink at the moment of release.

\subsection{Dynamic Modeling and Simulation}
To capture this parasitic energy loss, the energy balance was extended with rotational and dissipative terms:
\begin{equation}
\frac{1}{2} \sum_{i=1}^{4} k_i \left(\delta_{0,i}^2 - \delta_{r,i}^2\right) = \frac{1}{2} m v_{\text{CoM}}^2 + \frac{1}{2} I_{yy} \omega_p^2 + E_{\text{loss}}
\end{equation}
where $I_{yy}$ is the pitch moment of inertia and $\omega_p$ is the pitching rate at separation. To model the contact physics and transient impact forces, the system was simulated with the MuJoCo multi-body physics engine~\cite{todorov2012mujoco}.

In the simulation, the rover, including its J-legs, is modeled as a single rigid body; the energy absorbed by the J-legs is not modeled through explicit leg flexibility but is lumped into an increased equivalent damping of the chassis, tuned so that the simulated separation velocity matches the velocities measured in the early separation trials. With this calibration, the simulation indicated that the ejection springs must deliver substantially more energy than the rigid-body estimate to reach the required separation velocity. Based on the simulation results, the spring parameters (wire diameter, coil density, free length, and solid length) were recalculated and a stiffer set of custom springs was manufactured.

\subsection{Parametric Spring Rate Optimization and Vibration Failures}
The updated stiffer springs, installed in the Iteration 4 assembly, were subjected to the first vibration test campaign at the T\"{U}B\.{I}TAK UZAY satellite testing facilities. During Z-axis random vibration testing, the system failed. Post-test inspection revealed that the stored elastic energy was unbalanced across the four-spring array. This energy imbalance, coupled with manufacturing tolerances, caused one of the springs to jam within its guide housing, generating excessive dynamic bouncing (displacement) that halted the shaker table.

To resolve this energy imbalance and guide stiction while maintaining the required pushing force, a parametric spring optimization was conducted in two stages, balancing the stiffness of the spring interface, the delivered energy, and the force margin at release.

\begin{figure*}[htbp]
\centering
\setlength{\tabcolsep}{2pt}
\begin{tabular}{cc}
\includegraphics[width=0.48\textwidth]{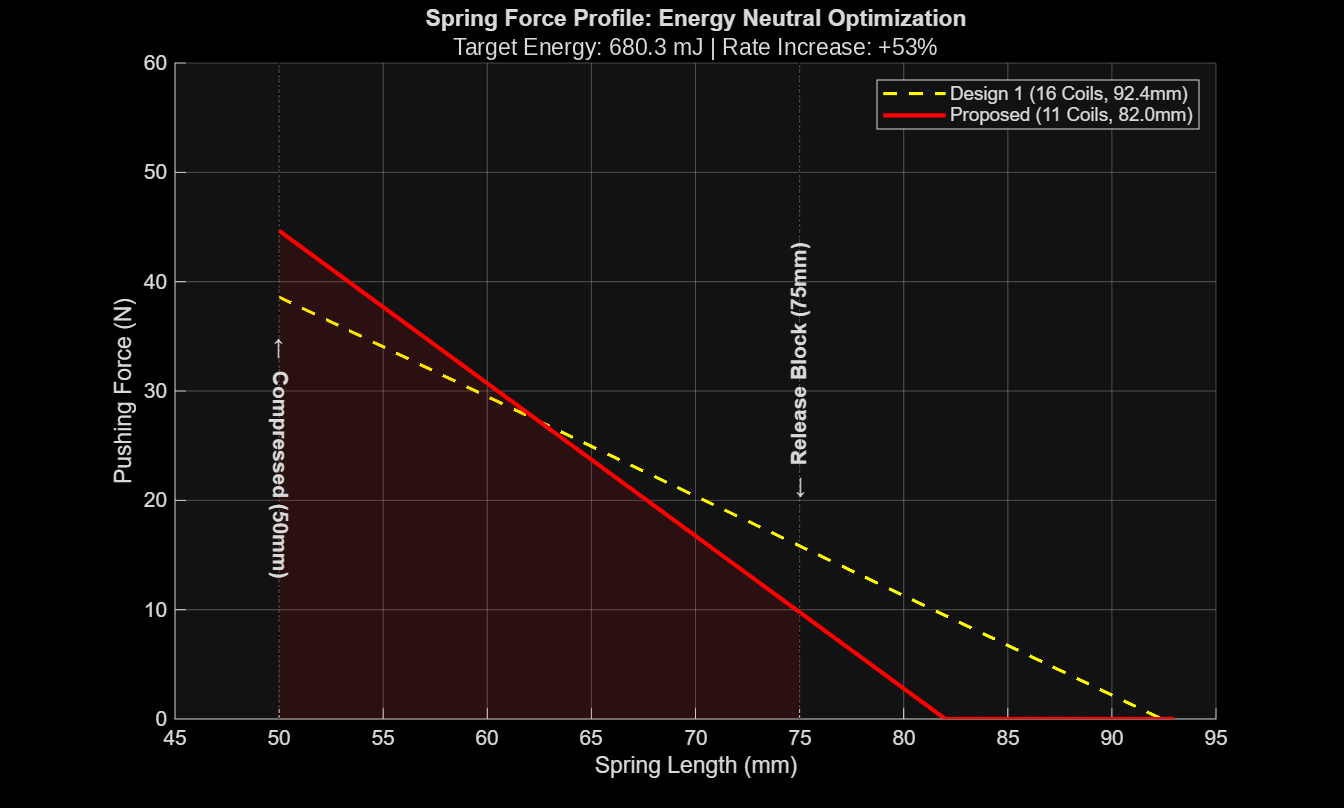} &
\includegraphics[width=0.45\textwidth]{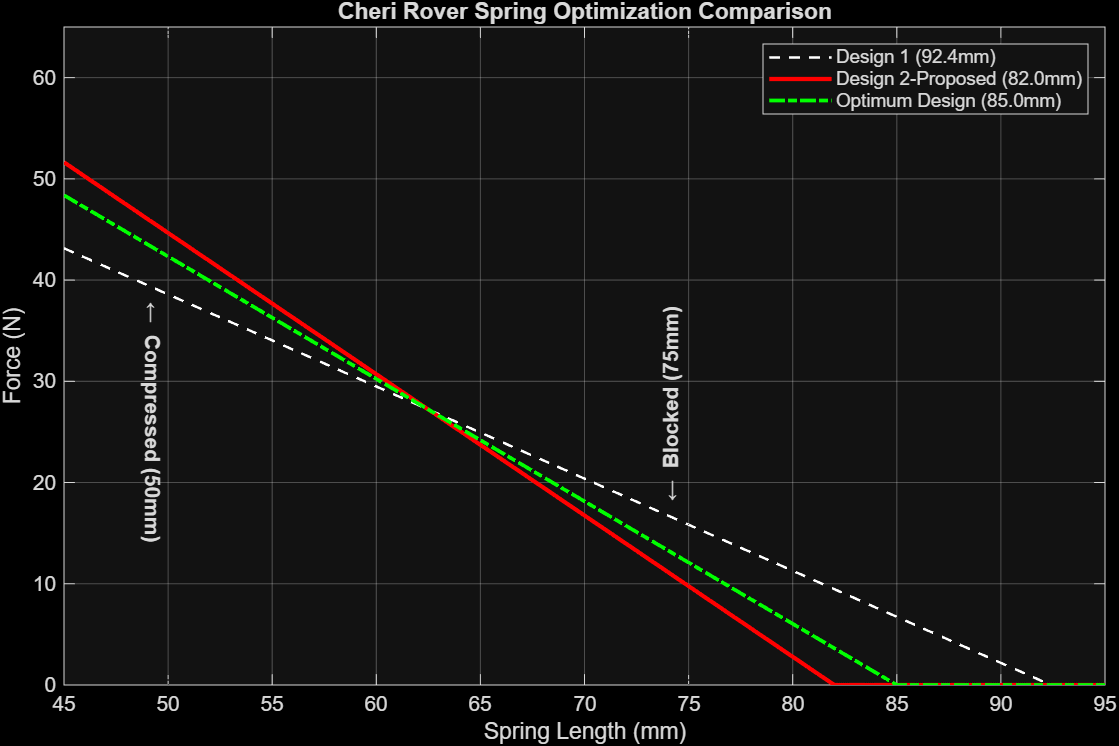} \\
(a) Spring-rate shift at constant delivered energy & (b) Force--length trade-off of the candidate designs
\end{tabular}
\caption{Two-stage parametric spring optimization: (a) increasing the spring rate by $+53\%$ while keeping the energy delivered per spring over the separation stroke constant at $680.3\,\text{mJ}$; (b) increasing the free length from $82.0\,\text{mm}$ (Proposed) to $85.0\,\text{mm}$ (Optimum), which raises the spring force at the release point ($75.0\,\text{mm}$) from $\approx 9.8\,\text{N}$ to $\approx 12.1\,\text{N}$ with a similar stowed force.}
\label{fig:parametric_spring_optimization_stages}
\end{figure*}

\subsubsection{Stage 1: Energy-Neutral Stiffness Increase}
First, to stiffen the spring-supported interface and reduce the dynamic displacement that caused jamming, the spring geometry was modified to obtain a steeper force profile without changing the energy delivered to the rover. The energy delivered by one linear spring between the stowed length ($50.0\,\text{mm}$) and the release point ($75.0\,\text{mm}$) is:
\begin{equation}
E = \frac{1}{2} k \left(\Delta L_0^2 - \Delta L_r^2\right) = 680.3\,\text{mJ}
\end{equation}
where $\Delta L_0 = L_f - 50.0\,\text{mm}$ and $\Delta L_r = L_f - 75.0\,\text{mm}$. Keeping this energy constant, the total coil count was reduced from $16.0$ to $11.1$ and the free length $L_f$ was shortened from $92.4\,\text{mm}$ to $82.0\,\text{mm}$. As plotted in Fig.~\ref{fig:parametric_spring_optimization_stages}(a), the spring rate $k$ increased from $0.910\,\text{N/mm}$ (Original) to $1.396\,\text{N/mm}$ (Proposed), a $+53.4\%$ increase.

\subsubsection{Stage 2: Stroke Optimization and Release Force Margin}
Although the Proposed ($82.0\,\text{mm}$) design increased the interface stiffness, its shorter free length leaves little residual compression at the release point. As the rover moves away from the ground plate, the spring force decays rapidly, leaving a small pushing force at release and increasing the risk of stiction-induced jamming.

To address this trade-off, a second stage increased the free length while keeping the stowed force and the delivered energy nearly unchanged. This led to the selection of the Optimized ($85.0\,\text{mm}$) spring configuration. The physical trade-offs are plotted in Fig.~\ref{fig:parametric_spring_optimization_stages}(b) and summarized in Table~\ref{tab:spring_parameters}:
\begin{itemize}
    \item \textbf{Similar Stowed Force:} At the stowed length ($50.0\,\text{mm}$, i.e., $35.0\,\text{mm}$ compression for the $85.0\,\text{mm}$ spring and $32.0\,\text{mm}$ for the $82.0\,\text{mm}$ spring), the Optimized and Proposed designs exert $\approx 42.3\,\text{N}$ and $\approx 44.7\,\text{N}$ per spring, respectively, well within the load limits of the central separation nut.
    \item \textbf{Longer Residual Compression:} The Optimized ($85.0\,\text{mm}$) spring remains compressed by $10.0\,\text{mm}$ at the release point, compared with $7.0\,\text{mm}$ for the Proposed spring, so that the spring is still pushing firmly when contact is lost. The delivered energy over the $25.0\,\text{mm}$ separation stroke is the same ($680.3\,\text{mJ}$) for both designs.
    \item \textbf{Force Margin at Release:} At the separation point ($75.0\,\text{mm}$), the Proposed spring force drops to $\approx 9.8\,\text{N}$, whereas the Optimized spring still delivers $\approx 12.1\,\text{N}$ (the Original spring delivered $\approx 15.8\,\text{N}$ at this point, but with the lower rate). The additional $\approx 2.3\,\text{N}$ per spring ($\approx 9\,\text{N}$ for the four-spring array) provides margin against transient friction or stiction within the guide housings.
\end{itemize}

\begin{table}[htbp]
\centering
\caption{Geometric and physical comparison of the evolutionary compression spring iterations.}
\label{tab:spring_parameters}
\footnotesize
\setlength{\tabcolsep}{2.5pt}
\begin{tabular}{lcccc}
\toprule
\textbf{Config.} & \textbf{Free length} & \textbf{Rate} & \textbf{Coils} & \textbf{Rate incr.} \\
 & $L_f$ (mm) & $k$ (N/mm) & $n_t$ & (\%) \\
\midrule
Original & 92.4 & 0.910 & 16.0 & -- \\
Proposed (82) & 82.0 & 1.396 & 11.1 & +53.4\% \\
\textbf{Optimized (85)} & \textbf{85.0} & \textbf{1.209} & \textbf{12.5} & \textbf{+32.9\%} \\
\bottomrule
\end{tabular}
\end{table}

Extensive laboratory release trials were performed to compare the three sets. The Optimized ($85.0\,\text{mm}$) design exhibited the most balanced energy release, the most repeatable pitching behavior, and separated without jamming in all trials. Consequently, this configuration was selected for the second environmental testing campaign at the T\"{U}B\.{I}TAK UZAY facilities.

During the second vibration test campaign, the optimized springs and guides (Iteration 5) showed no jamming or bouncing. However, the assembly failed the test because its first natural frequency was measured at $34\,\text{Hz}$, well below the $70\,\text{Hz}$ lander requirement. Since the spring guides were functioning nominally, this low-frequency resonance was traced directly to the structural flexing of the flat aluminum ground plate itself, creating a clear engineering requirement for structural topology optimization, which is detailed in Section~\ref{sec:structural_dynamics}.

\section{Structural Dynamics and Topology Optimization of the Ground Plate}
\label{sec:structural_dynamics}

To survive the launch vibration environment without structural failure or premature separation, the stowed micro-rover assembly must satisfy stringent dynamic stiffness requirements. This section details the vibration test at the T\"{U}B\.{I}TAK UZAY satellite testing facilities that revealed the low first mode of the flat ground plate, the Solid Isotropic Material with Penalization (SIMP) topology optimization used to redesign the plate, the comparative finite element modal analyses of the two designs, and the vibration and shock tests of the final $84\,\text{Hz}$ configuration.

\subsection{The 34\,Hz Resonance (Iteration 5 Failure)}
The fifth-generation separation system prototype, featuring the tribologically-optimized guide housings, custom-designed springs, and a flat $5.0\,\text{mm}$ thick aluminum 7075-T6 ground plate, was subjected to the second vibration test campaign on an electrodynamic shaker system at the satellite testing facility of T\"{U}B\.{I}TAK UZAY. The test sequence comprised sinusoidal and random vibration at the levels specified for the lunar lander, with the $5.0\,\text{kg}$ rover model stowed and preloaded on the plate.

During the initial low-level sine sweep in the Z-axis (normal to the plate), the first natural frequency of the stowed assembly was measured at $34\,\text{Hz}$, far below the $f_1 \ge 70\,\text{Hz}$ requirement of the lander specification~\cite{bice2025devspec}, which keeps the payload modes clear of the primary structural modes of the spacecraft and launch vehicle. At $34\,\text{Hz}$, the flat ground plate exhibited large out-of-plane bending and torsional deformation, amplifying the displacement of the stowed rover body. Continuing to full level would have risked overloading the central separation nut~\cite{sfr_datasheet} and damaging the rover's structural panels, so the test was stopped and recorded as a failure.

A post-test finite element analysis (FEA) showed that the low natural frequency was not caused by compliance in the spring guides or the actuator assembly themselves. Instead, the heavier guide housings and spring sets of Iteration 5, mounted on the same flat plate, lowered the first natural frequency from the $51\,\text{Hz}$ measured for Iteration 4, and the mostly flat, unreinforced ground plate deformed as a flexible membrane. To raise the natural frequency to $f_1 \ge 70\,\text{Hz}$ without violating the newly allocated mass limit of $1.6\,\text{kg}$ (relaxed from the initial $1.2\,\text{kg}$ during the separation-nut era), a redistribution of material by structural topology optimization was required.

\subsection{Mathematical Formulation of SIMP Topology Optimization}
To maximize the fundamental natural frequency of the ground plate while satisfying the mass budget and manufacturability constraints, a Solid Isotropic Material with Penalization (SIMP) topology optimization~\cite{bendsoe1989optimal} was formulated. The optimization domain $\Omega$ was defined based on the envelope of the ground plate, with non-design spaces reserved around the six M5 mounting bolt holes, the four guide housings, and the central separation nut interface.

The design variable is the continuous relative density field $x(\mathbf{y}) \in [0, 1]$ defined at each point $\mathbf{y} \in \Omega$. The goal is to maximize the first eigenvalue $\lambda_1$ (corresponding to the square of the first fundamental natural frequency $\omega_1^2$) of the stowed assembly. The optimization problem is mathematically formulated as:

\begin{equation}
\label{eq:simp_opt}
\begin{aligned}
    & \underset{x}{\text{maximize}} && \lambda_1(\mathbf{x}) = \omega_1^2 \\
    & \text{subject to} && \mathbf{K}(\mathbf{x}) \mathbf{\Phi}_r = \lambda_r \mathbf{M}(\mathbf{x}) \mathbf{\Phi}_r, \quad r=1, 2, \dots \\
    & && V(\mathbf{x}) = \int_{\Omega} x(\mathbf{y})\,d\Omega \le V^* \\
    & && 0 < x_{\min} \le x(\mathbf{y}) \le 1, \quad \forall \mathbf{y} \in \Omega
\end{aligned}
\end{equation}

where $\mathbf{K}(\mathbf{x})$ is the global stiffness matrix, $\mathbf{M}(\mathbf{x})$ is the global mass matrix, $\mathbf{\Phi}_r$ is the $r$-th eigenvector (mode shape), $V^*$ is the target volume corresponding to the allowable mass limit, and $x_{\min} = 10^{-3}$ is a small positive lower bound introduced to avoid singularity of the stiffness matrix in void regions.

Under the SIMP method, the material properties of each element are parameterized through the penalization of the relative density. The Young's modulus $E_i$ and material density $\rho_i$ of each element $i$ are interpolated as:
\begin{equation}
    E_i(x_i) = E_{\min} + x_i^p (E_0 - E_{\min})
\end{equation}
\begin{equation}
    \rho_i(x_i) = x_i \rho_0
\end{equation}
where $E_0$ is the nominal Young's modulus of Al 7075-T6 ($71.7\,\text{GPa}$), $E_{\min}$ is the stiffness of the void material ($E_0 \times 10^{-9}$), $\rho_0$ is the nominal density of the alloy ($2810\,\text{kg/m}^3$), and $p$ is the penalization power. To force the continuous density field toward a binary solid-void ($0$--$1$) distribution suitable for computer numerical control (CNC) milling, the penalization power was set to $p = 3$, which satisfies the Hashin--Shtrikman bounds for a Poisson's ratio of $1/3$~\cite{bendsoe1999material}.

Because the stowed $5.0\,\text{kg}$ rover represents a significant portion of the total vibrating mass, the global mass matrix $\mathbf{M}(\mathbf{x})$ was formulated as:
\begin{equation}
    \mathbf{M}(\mathbf{x}) = \mathbf{M}_{\text{plate}}(\mathbf{x}) + \mathbf{M}_{\text{rover}} + \mathbf{M}_{\text{components}}
\end{equation}
where $\mathbf{M}_{\text{plate}}$ is the design-dependent mass matrix of the ground plate, $\mathbf{M}_{\text{components}}$ represents the non-structural localized masses of the spring guides and actuator ($0.26\,\text{kg}$) and of the four separation spring sets, each comprising a spring, guide bushing and guide rod ($0.43\,\text{kg}$ in total), and $\mathbf{M}_{\text{rover}}$ is the inertia of the stowed rover, modeled as a concentrated mass of $5.0\,\text{kg}$ at the rover's Center of Mass (CoM). This concentrated mass was coupled to the plate via rigid kinematic constraints representing the preloaded contact interfaces of the four J-legs and the central separation nut.

The optimization was carried out with the topology optimization module of ANSYS Mechanical and solved iteratively using the Method of Moving Asymptotes (MMA)~\cite{svanberg1987method}. To prevent checkerboard patterns and establish a minimum manufacturable feature size, a sensitivity filter~\cite{sigmund1998numerical} with a filter radius of $r_{\min} = 6.0\,\text{mm}$ was applied to the density derivatives.

\subsection{Comparative Modal and Finite Element Analysis (FEA)}
The resulting optimal material distribution from the SIMP optimization was post-processed and converted into a CAD model by pocketing out the low-stress regions and generating a network of diagonal stiffening ribs. The thickness of the main plate was reduced in non-structural regions to $1.5\,\text{mm}$, while the stiffening ribs and mounting perimeters maintained a thickness of up to $8.0\,\text{mm}$. The final mass of the optimized sixth-generation ground plate was $0.85\,\text{kg}$; together with the spring guides and actuator ($0.26\,\text{kg}$) and the four separation spring sets ($0.43\,\text{kg}$), the total separation assembly mass is $1.54\,\text{kg}$, within the $1.6\,\text{kg}$ budget.

To verify the design before manufacturing, comparative finite element modal analyses were conducted for both the flat fifth-generation plate (Iteration 5) and the topology-optimized sixth-generation plate (Iteration 6). The models were built from the hardware geometry and nominal material properties and were not calibrated against the vibration test data. Both models were simulated under identical boundary conditions:
\begin{enumerate}
    \item \textbf{Mounting Interfaces:} Fixed boundary constraints were applied at the six M5 bolt mounting holes.
    \item \textbf{Thermal Isolation Pads:} The $5.0\,\text{mm}$ thick polyimide thermal isolation pads located between the ground plate and the lander mating surface were modeled as elastic spring contacts with a combined axial stiffness of $k_{\text{axial}} = 1.2 \times 10^9\,\text{N/m}$ for the six pads (approximately $2 \times 10^8\,\text{N/m}$ each), representing the compressive elasticity of the polymer under preload.
    \item \textbf{Preload Effects:} A bolt pre-tension load of $4.5\,\text{kN}$ was applied to the central separation nut to simulate the stowed preloading state.
\end{enumerate}

\begin{figure*}[htbp]
\centering
\setlength{\tabcolsep}{2pt}
\begin{tabular}{cc}
\includegraphics[width=0.47\textwidth]{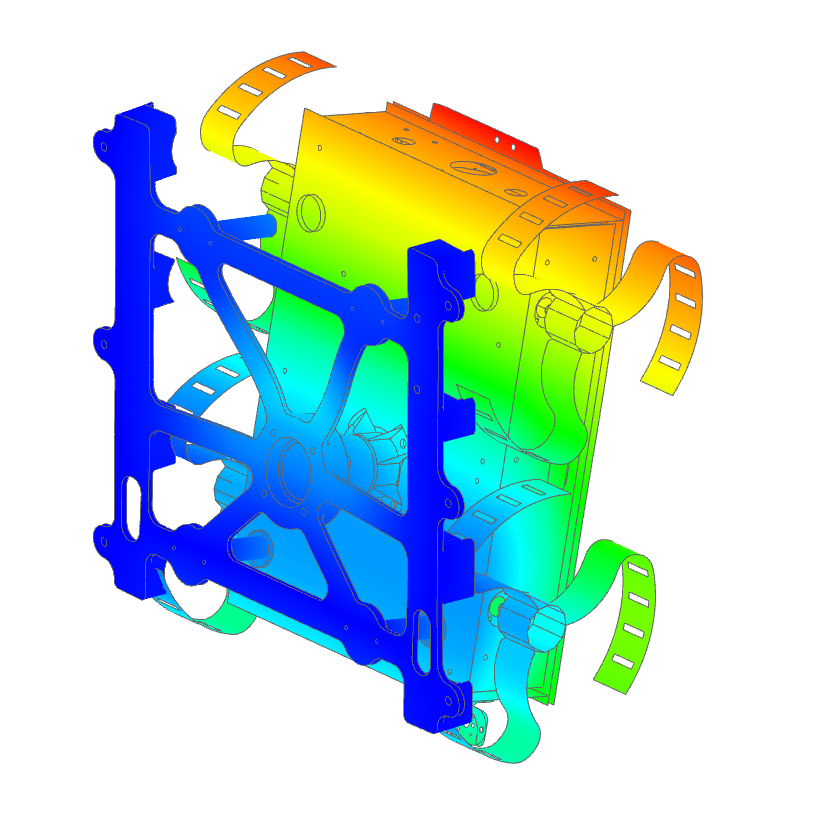} &
\includegraphics[width=0.47\textwidth]{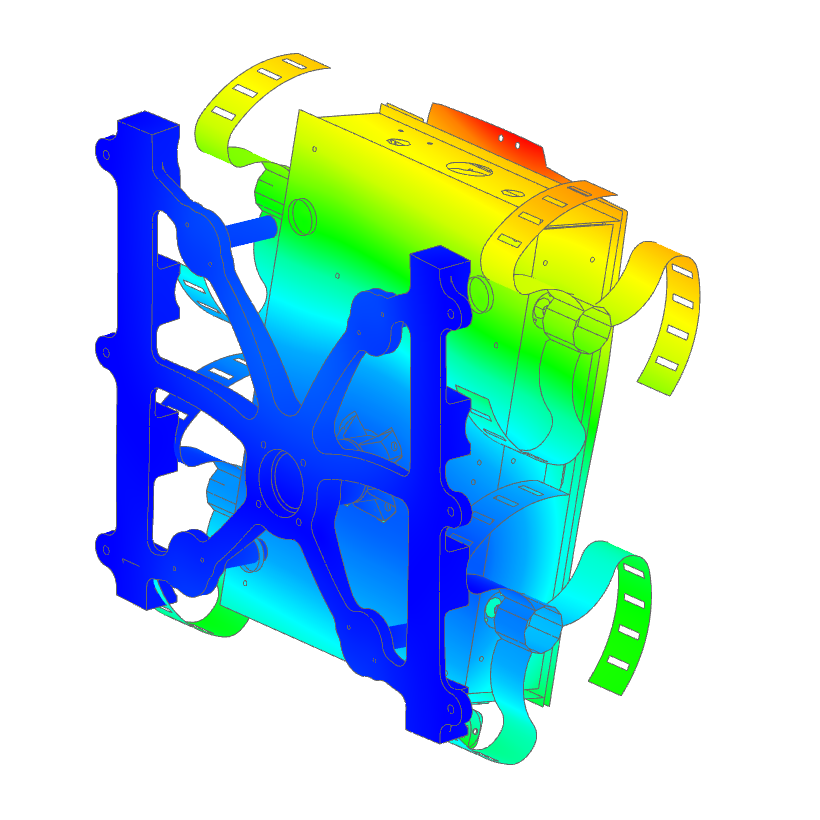} \\
(a) Iteration 5: Flat Plate ($34\,\text{Hz}$ First Mode) & (b) Iteration 6: Optimized Plate ($84\,\text{Hz}$ First Mode)
\end{tabular}
\caption{Finite element modal analysis comparison showing the first fundamental mode shapes of (a) the flat fifth-generation ground plate exhibiting severe central flexing at $34\,\text{Hz}$, and (b) the topology-optimized sixth-generation ground plate with stiffening ribs raising the natural frequency to $84\,\text{Hz}$.}
\label{fig:modal_comparison}
\end{figure*}

\begin{table}[htbp]
\centering
\caption{First three natural frequencies of the stowed separation assembly before and after topology optimization (FEA). Modes are matched by deformation shape.}
\label{tab:modal_frequencies}
\footnotesize
\setlength{\tabcolsep}{3.5pt}
\begin{tabular}{clccc}
\toprule
\textbf{Mode} & \textbf{Deformation Shape} & \textbf{Pre-Opt.} & \textbf{Post-Opt.} & \textbf{Status} \\
\textbf{No.} & & \textbf{Iter. 5 (Hz)} & \textbf{Iter. 6 (Hz)} & \textbf{($\ge 70\,\text{Hz}$)} \\
\midrule
1 & Longitudinal Torsion & 33.6 & 84.0 & Passed \\
2 & Out-of-plane Bending & 55.4 & 116.9 & Passed \\
3 & Lateral Torsion      & 59.8 & 107.2 & Passed \\
\bottomrule
\end{tabular}
\end{table}

The numerical results are plotted in Fig.~\ref{fig:modal_comparison} and summarized in Table~\ref{tab:modal_frequencies}. The first mode shape of the flat plate (Fig.~\ref{fig:modal_comparison}(a)) is dominated by longitudinal twisting centered near the rover CoM. The asymmetric spring placement adds a bending moment at the plate connection, coupling this twist with out-of-plane bending so that the rover CoM oscillates with large amplitude. This mode occurs at $33.6\,\text{Hz}$ (measured: $34\,\text{Hz}$), within the band of the low-frequency launch excitation.

In contrast, in the optimized plate (Fig.~\ref{fig:modal_comparison}(b)) the strain energy of the first mode is carried mainly by the diagonal stiffening ribs. The first natural frequency is raised to $84.0\,\text{Hz}$ (Table~\ref{tab:modal_frequencies}), $+150\%$ relative to the computed $33.6\,\text{Hz}$ of Iteration 5 and $20\%$ above the $70\,\text{Hz}$ lander requirement. The second mode (out-of-plane bending) is pushed from $55.4\,\text{Hz}$ to $116.9\,\text{Hz}$, while the third mode (lateral torsion) rises from $59.8\,\text{Hz}$ to $107.2\,\text{Hz}$. Note that, after optimization, the lateral torsion mode ($107.2\,\text{Hz}$) lies below the out-of-plane bending mode ($116.9\,\text{Hz}$); modes in Table~\ref{tab:modal_frequencies} are matched by shape rather than by order. These analyses indicated that the sixth-generation design would satisfy the frequency requirement in shaker testing.

\subsection{Vibration and Shock Test Setups}
\begin{figure*}[htbp]
\centering
\setlength{\tabcolsep}{2pt}
\begin{tabular}{ccc}
\includegraphics[width=0.31\textwidth]{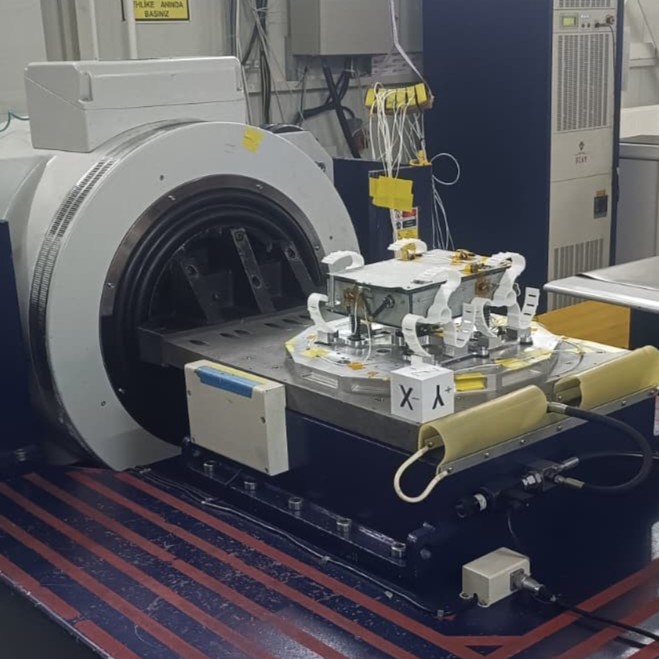} &
\includegraphics[width=0.31\textwidth]{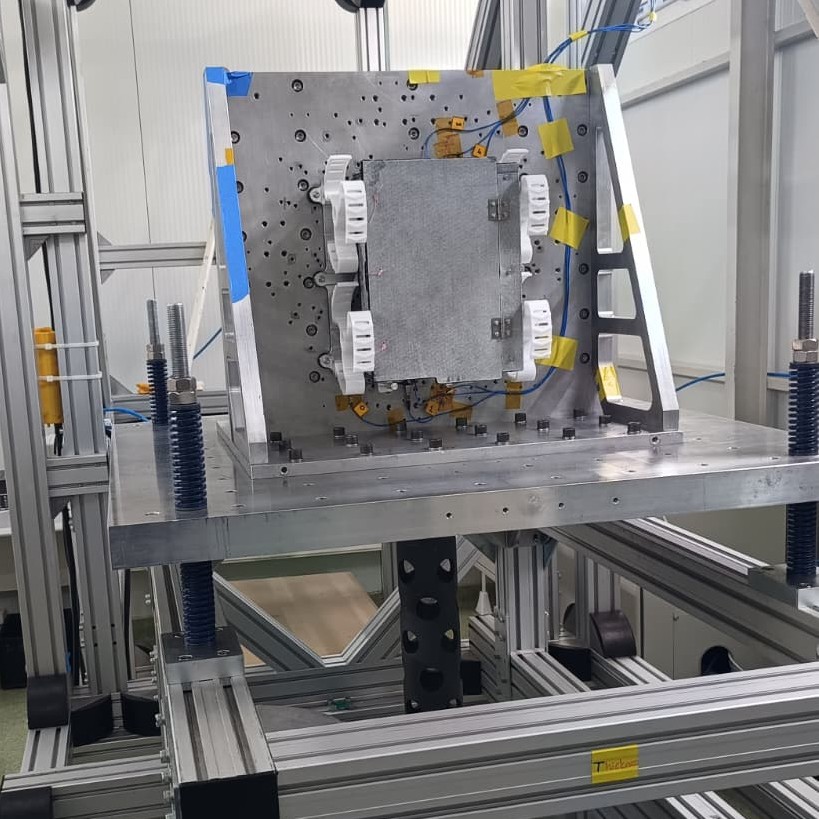} &
\includegraphics[width=0.31\textwidth]{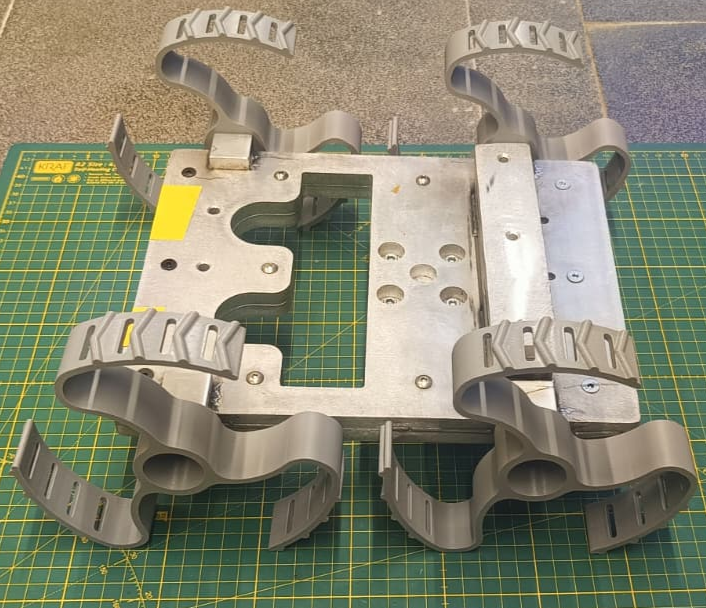} \\
(a) Vibration test setup on shaker table & (b) Pneumatic shock test setup & (c) Dummy model (mass simulator)
\end{tabular}
\caption{Structural test and calibration setups at the T\"{U}B\.{I}TAK UZAY satellite testing facility: (a) the stowed separation assembly mounted on the electrodynamic shaker table for vibration testing in three orthogonal axes, (b) the pneumatic shock test setup, in which an aluminum projectile strikes a resonant plate, and (c) the non-functional dummy model used as a mass simulator.}
\label{fig:environmental_test_setups}
\end{figure*}

To evaluate the separation assembly in environmental testing without risking the structural model during test set-up, a two-model strategy was used: a non-functional dummy model (shown in Fig.~\ref{fig:environmental_test_setups}(c)) for preliminary system checks and calibration, and a structurally representative Structural Model (SM) for the test runs at the satellite testing facilities of T\"{U}B\.{I}TAK UZAY. The physical configurations of these setups are illustrated in Fig.~\ref{fig:environmental_test_setups}.

The dummy model played a vital role in protecting the structural model during the setup phases of both campaigns. It was engineered to replicate only the gross geometric envelope, the total $5.0\,\text{kg}$ mass, and the core bolt patterns of the rover, making it suitable for preliminary checks:
\begin{itemize}
    \item \textbf{Vibration Test System Check:} In the vibration campaign, the dummy model was mounted first solely to perform a comprehensive system check. It was used to verify fixture stiffness, mounting clearances, and accelerometer wiring, and to run low-level diagnostic sine sweeps before mounting the SM.
    \item \textbf{Impact Parameter Tuning:} In the shock campaign, the dummy model was used to tune the pneumatic shock simulator. Because the resulting shock response spectrum (SRS) is sensitive to the set-up, the compressed-air pressure (up to $4\,\text{bar}$), projectile travel, resonant plate, and damping pads were adjusted with the dummy until the specified SRS was achieved, before the SM was installed.
\end{itemize}

Once the set-up checks were complete, the test runs were executed with the Structural Model (SM), which replicates the flight rover's $5.0\,\text{kg}$ mass, Center of Mass (CoM) location, mass moments of inertia, structural panel stiffness, and J-leg stiffness.

The physical vibration test setup was established using the SM. The fully assembled sixth-generation topology-optimized separation system---comprising the optimized ground plate (Iteration 6), custom guide housings, Optimized ($85.0\,\text{mm}$) springs, and the stowed $5.0\,\text{kg}$ SM preloaded to $4.5\,\text{kN}$ via the central separation nut---was rigidly bolted to a heavy, custom fixture on the sliding table of an electrodynamic shaker system. Triaxial piezoelectric accelerometers were installed at positions selected from the finite element mode shapes. The assembly was then tested in all three orthogonal axes at the specified levels: sinusoidal vibration over $5$--$100\,\text{Hz}$ at $4\,\text{oct/min}$ (up to $6\,\text{g}$ normal to the mounting surface and $5\,\text{g}$ in-plane), and random vibration over $20$--$2000\,\text{Hz}$ ($0.050\,\text{g}^2/\text{Hz}$ plateau, $9.8\,\text{g}_{\text{rms}}$, $1\,\text{min}$ per axis). These are the acceptance-level inputs; the qualification level defined in the lander specification~\cite{bice2025devspec} is $9\,\text{g}$ sine and $15.5\,\text{g}_{\text{rms}}$ random for $2\,\text{min}$ per axis.

Before and after each sweep, low-level sinusoidal sweeps ($5$--$2200\,\text{Hz}$ at $0.5\,\text{g}$) were performed to search for resonances and identify natural frequencies through frequency responses, and compare the change in system behavior as the tests progressed to evaluate possible sustained damage. A shift of $\le 5\%$ in the resonant frequencies between the pre- and post-test sweeps was taken as the pass criterion. The optimized plate passed the vibration tests without detectable structural degradation, and the first resonance was measured at $84\,\text{Hz}$, above the $70\,\text{Hz}$ lander requirement.

Following the vibration campaign, the assembly was subjected to shock testing using the SM to verify structural integrity under the pyrotechnic separation shocks expected during the mission. The shock was generated with a pneumatic shock simulator in accordance with ECSS-E-ST-10-03C~\cite{ecss2012testing}, with a tolerance of $-3/+6\,\text{dB}$ on the SRS. The system accelerates an aluminum projectile with compressed air (up to $4\,\text{bar}$) to strike a resonant plate. The facility can generate up to $3000\,\text{g}$ in the time domain and up to $8000\,\text{g}$ in the SRS; the specified test level was an SRS ($Q = 10$) rising at $+9\,\text{dB/oct}$ from $38.4\,\text{g}$ at $100\,\text{Hz}$ to $2200\,\text{g}$ at $1500\,\text{Hz}$ and constant at $2200\,\text{g}$ up to $4000\,\text{Hz}$, applied once in each of the three axes, as specified for acceptance testing (qualification requires three shocks per axis~\cite{bice2025devspec}).

Shock accelerometers (range up to $5000\,\text{g}$) at the interface points recorded the transient response. The SM provided the representative load path to verify that the central separation nut and the spring guide housings did not suffer premature release, fracture, or plastic deformation. No premature release or structural damage was observed after the shock tests.

\section{Experimental Verification and Trajectory Validation}
\label{sec:experimental_validation}

To validate the analytical spring-rate optimization of
Section~\ref{sec:analytical_optimization} and to establish the separation
performance of the rover over a range of terrain slopes, a combined numerical
and experimental campaign was conducted. The campaign is structured around a
deliberate division of roles: a physical, gravity-scaled laboratory test is used
to \emph{validate the numerical model} under conditions that can be reproduced
on Earth, and the validated model is then used to \emph{predict} flight
behavior under lunar gravity, which the laboratory cannot reproduce. This
section sets out the scaling methodology and its limits, the motion-capture
instrumentation, the slope test matrix, and the comparison between the two.

\subsection{Gravity-Scaled Testing Methodology and Its Limits}
\label{sec:gravity_scaling}

Reproducing the lunar surface gravity of
$g_{\text{Moon}} = 1.62\,\text{m/s}^2$ on Earth is the central difficulty in
validating a separation system. Active compensation using tensioned suspension
wires or overhead gantries introduces parasitic forces, cable inertia and
mechanical friction that corrupt the millisecond-scale separation transient of a
lightweight micro-rover. A passive alternative was therefore adopted: a
dedicated $1/6$ mass model was manufactured, so that its weight under terrestrial
gravity matches the weight of the flight article under lunar gravity.

The weight of the $5.0\,\text{kg}$ flight rover on the Moon is
\begin{equation}
F_{g,\text{Moon}} = m_{\text{flight}}\, g_{\text{Moon}}
= 5.0\,\text{kg} \times 1.62\,\text{m/s}^2 \approx 8.10\,\text{N},
\end{equation}
and the weight of the scaled model under terrestrial gravity is
\begin{equation}
F_{g,\text{sim}} = m_{\text{sim}}\, g_0
= 0.833\,\text{kg} \times 9.81\,\text{m/s}^2 \approx 8.17\,\text{N}
\approx F_{g,\text{Moon}} .
\end{equation}
The residual $0.9\%$ discrepancy arises because
$g_{\text{Moon}}/g_0 = 1/6.06$ rather than exactly $1/6$; it is within the
tolerance of the spring preload and is neglected hereafter.

It must be stated explicitly that matching weight does \emph{not} reproduce
lunar gravity. Weight and inertia scale together with mass, so the ratio
$F/m = g$ is unchanged and the scaled model still falls at
$9.81\,\text{m/s}^2$. Equivalently, the free-flight trajectory
\begin{equation}
z(y) = h - \frac{g}{2}\left(\frac{y}{v_{\text{sep}}}\right)^{2}
\label{eq:trajectory}
\end{equation}
contains no mass term whatsoever; only the gravitational acceleration $g$ and
the separation velocity $v_{\text{sep}}$ appear. The mass reduction acts solely
through $v_{\text{sep}}$.

Geometric similarity of the free-flight phase is nevertheless obtained, and
obtained automatically. Requiring that Eq.~\eqref{eq:trajectory} describe the
same curve in both environments at a common release height $h$ gives the Froude
condition $g/v_{\text{sep}}^{2} = \text{const}$, hence
\begin{equation}
v_{\text{sep,Earth}} = v_{\text{sep,Moon}}\sqrt{\frac{g_0}{g_{\text{Moon}}}}
\approx 2.46\, v_{\text{sep,Moon}} .
\label{eq:froude_velocity}
\end{equation}
A compression spring is an energy source rather than an impulse source: it
delivers the energy $E$ stored over its stroke (Section~\ref{sec:analytical_optimization}) irrespective of the mass
it acts against, so that
\begin{equation}
v_{\text{sep}} = \sqrt{\frac{2E}{m}}
\quad\Longrightarrow\quad
\frac{v_{\text{sep,Earth}}}{v_{\text{sep,Moon}}}
= \sqrt{\frac{m_{\text{flight}}}{m_{\text{sim}}}} = \sqrt{6} \approx 2.45.
\label{eq:spring_energy}
\end{equation}
This argument assumes that the fraction of the spring energy converted into
translational kinetic energy is the same for both masses. As quantified
below, that fraction is only about
$20$--$55\%$, because most of the energy is absorbed by J-leg deformation,
structural damping, rotation and friction; its independence of mass is
therefore an assumption of the scaling rather than a measured result.
Equations~\eqref{eq:froude_velocity} and \eqref{eq:spring_energy} coincide to within $0.5\%$.
The flight-specification springs therefore satisfy the Froude condition on the
$1/6$ mass model with \emph{no} re-tuning of stiffness or preload, and the
laboratory article traces the same spatial path and arrives at the surface with
the same attitude as the flight article. The flight time is shorter by
$\sqrt{6.06} = 2.46$. Aerodynamic drag, absent on the Moon, is neglected in this
argument; for the light model it shortens the horizontal travel over the
$0.6\,\text{s}$ laboratory flight by an estimated $1$--$3\%$.

Similarity is confined to the free-flight phase. Table~\ref{tab:scaling} lists
the quantities that transfer and those that do not. The distinction governs the
interpretation of every result in this section.

\begin{table}[htbp]
\centering
\caption{Ratio of the terrestrial $1/6$ mass test to lunar flight conditions
under weight-matched Froude scaling, with $f = m_{\text{sim}}/m_{\text{flight}}
= 1/6$ and $g_0/g_{\text{Moon}} = 6.06$. Quantities with unit ratio transfer
exactly; the remainder do not.}
\label{tab:scaling}
\footnotesize
\setlength{\tabcolsep}{3pt}
\begin{tabular}{lcc}
\toprule
Quantity & Scaling law & Ratio (Earth/Moon) \\
\midrule
Weight                        & $f\,(g_0/g_{\text{Moon}})$   & $1.01$ \\
Downrange at touchdown        & $f^{-1/2}(g_{\text{Moon}}/g_0)^{1/2}$ & $1.00$ \\
Attitude at touchdown         & $f^{-1/2}(g_{\text{Moon}}/g_0)^{1/2}$ & $1.00$ \\
Translational kinetic energy  & $f \cdot f^{-1}$             & $1.00$ \\
Rotational kinetic energy     & $f \cdot f^{-1}$             & $1.00$ \\
\midrule
Separation velocity           & $f^{-1/2}$                   & $2.45$ \\
Impact velocity               & $(g_0/g_{\text{Moon}})^{1/2}$ & $2.46$ \\
Angular rate                  & $f^{-1/2}$                   & $2.45$ \\
Flight time                   & $(g_{\text{Moon}}/g_0)^{1/2}$ & $0.41$ \\
Linear momentum               & $f^{1/2}$                    & $0.41$ \\
Contact duration              & $[f\,(g_{\text{Moon}}/g_0)]^{1/2}$ & $0.17$ \\
Soil frictional bearing term  & $g_0/g_{\text{Moon}}$        & $6.06$ \\
\bottomrule
\end{tabular}
\end{table}

The last entry is the most consequential and is easily overlooked. The
frictional term of the bearing capacity of a granular medium scales with the
product $\rho g$. The same regolith simulant tested under terrestrial gravity is
therefore approximately six times stiffer in bearing than it would be on the
Moon, while the rover's weight has been held constant by construction. For a
Bekker pressure--sinkage law~\cite{bekker1969introduction} $\sigma(z) = (k_c/b + k_\varphi)z^{n}$ with
$k_\varphi$ dominant, the static sinkage under an unchanged contact pressure is
reduced by a factor of approximately $6^{1/n}$. The laboratory surface consequently absorbs
substantially less energy during touchdown than lunar regolith would, and
returns a correspondingly larger fraction of the impact energy to the rover as
rebound and rotation. The terrestrial test is thus expected to be
\emph{conservative} with respect to landing stability. This effect is not
represented in the simulations of Section~\ref{sec:slope_campaign}, which use
the same contact parameters under both gravities, and it has not been confirmed
by measurement under lunar gravity.

\subsection{6-DOF Motion Tracking and Signal Processing}

\begin{figure}[htbp]
\centering
\includegraphics[width=0.48\textwidth]{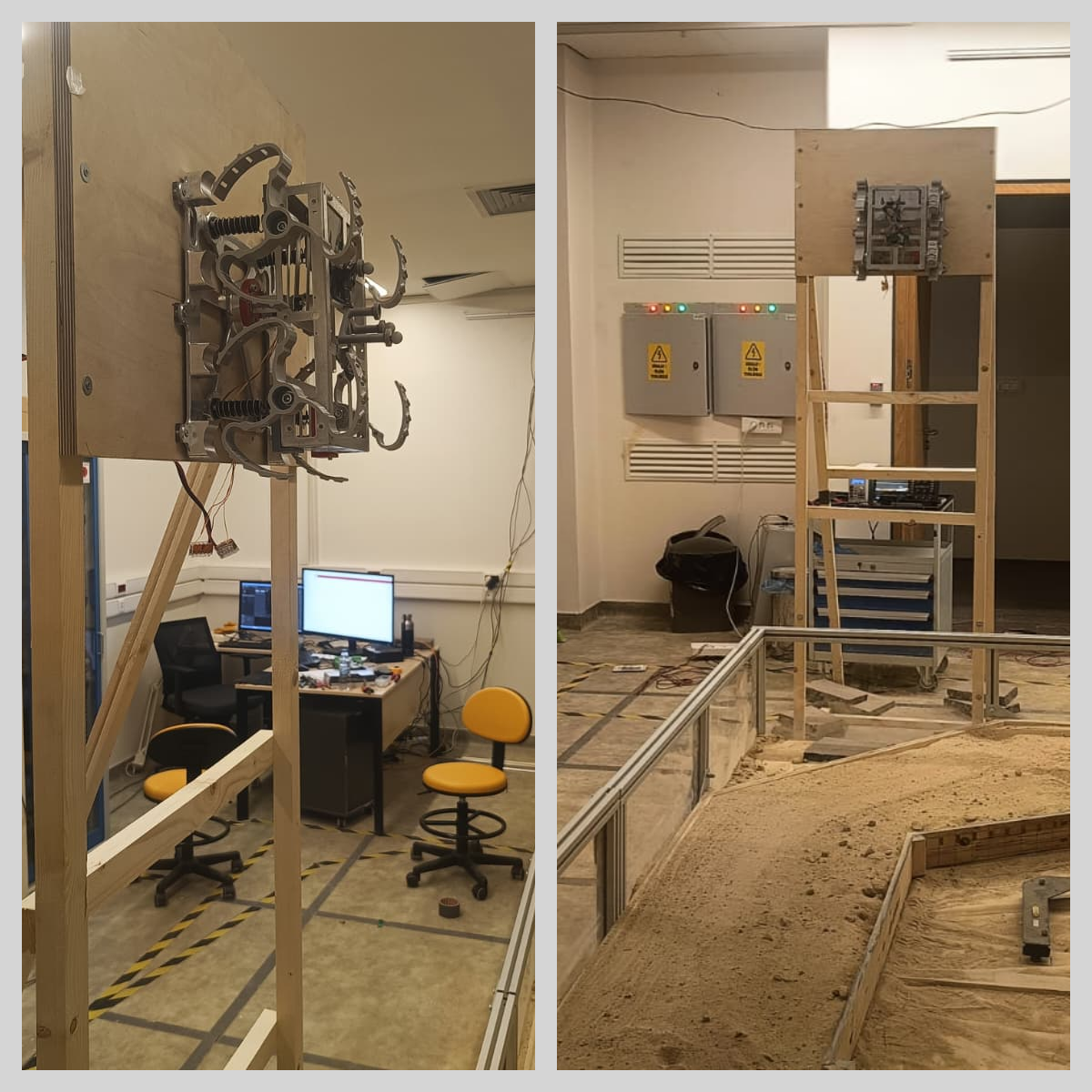}
\caption{Physical experimental launch test setup, showing the test arena with
wall-mounted optical motion-capture cameras and the landing sandbox (left), and
the $1/6$ mass model stowed and preloaded on the ground plate (right).}
\label{fig:launch_setup}
\end{figure}

The experimental launch rig consisted of the manufactured sixth-generation
topology-optimized ground plate fitted
with the optimized ($85.0\,\text{mm}$) ejection springs and the central
separation nut. The $1/6$ mass model was stowed and preloaded against the plate
at the flight mounting height, $H = 1.7\,\text{m}$ above the simulant surface.
The laboratory layout---the simulated lander platform, the stowed model, the
sandbox containing lunar regolith simulant and the surrounding motion-capture
array---is shown in Fig.~\ref{fig:launch_setup}.

An infrared motion-capture system was used to record the six-degree-of-freedom
rigid-body motion during the separation transient. Eight infrared cameras
operated at a sampling rate of $240\,\text{Hz}$ with a spatial resolution of
$0.1\,\text{mm}$. Four retro-reflective markers were placed symmetrically on the
model to define the Rover Coordinate System (RCS), whose origin was calibrated
to coincide with the model's center of mass. Camera calibration and marker
visibility trials were performed prior to the active runs using static and
low-velocity manual releases.

Raw marker data were resolved into the translational coordinates $(x,y,z)$ and
the Euler angles (pitch $\theta$, roll $\phi$, yaw $\psi$) of the rover relative
to the stationary Launcher Coordinate System (LCS). A zero-phase second-order
Butterworth low-pass filter with a cut-off frequency of $30\,\text{Hz}$ was
applied to suppress tracking jitter. Because the scaling compresses the separation transient by the factor $2.46$,
the effective temporal resolution referred to flight conditions is
$240/2.46 \approx 98\,\text{Hz}$ (approximately $19$ samples during the $0.08\,\text{s}$ spring stroke of the terrestrial test); this is adequate for the trajectory and
attitude histories reported here, but not for resolving the contact transient
itself.

\subsection{Numerical Model and Slope Test Campaign}
\label{sec:slope_campaign}

The multi-body dynamic model was implemented in MuJoCo. The rover, including its J-legs, was
represented as a single rigid body with mass and inertia tensor taken from the
flight CAD model; the energy absorbed by the flexible J-legs during the
separation stroke is represented by an increased equivalent damping of the
chassis (Section~\ref{sec:analytical_optimization}). Contact with the regolith was represented by a penalty formulation
whose normal stiffness was derived from Bekker pressure--sinkage parameters for
lunar mare regolith; the same contact parameters were used for both gravity
conditions. The separation springs were modeled as displacement-driven
elements, $F = k(x_0 - s)$, so that the delivered impulse is an outcome of the
accelerated mass rather than a prescribed input; this is necessary for the
scaling of Eq.~\eqref{eq:spring_energy} to be reproduced by the model rather
than imposed on it.

Three series were run across slopes of $0^\circ$, $\pm10^\circ$, $\pm20^\circ$,
$\pm30^\circ$ and $\pm45^\circ$, with positive denoting counter-clockwise and
negative clockwise inclination in the separation plane:

\begin{enumerate}
\item \textbf{Lunar simulation}: flight mass $5.0\,\text{kg}$,
      $g = 1.62\,\text{m/s}^2$.
\item \textbf{Terrestrial simulation}: $0.833\,\text{kg}$,
      $g = 9.81\,\text{m/s}^2$, identical spring specification.
\item \textbf{Terrestrial experiment}: the physical $1/6$ mass model landing in
      the sandbox, with the regolith simulant shaped to the target inclination,
      identical spring specification.
\end{enumerate}

\begin{figure}[htbp]
\centering
\includegraphics[width=1.0\columnwidth]{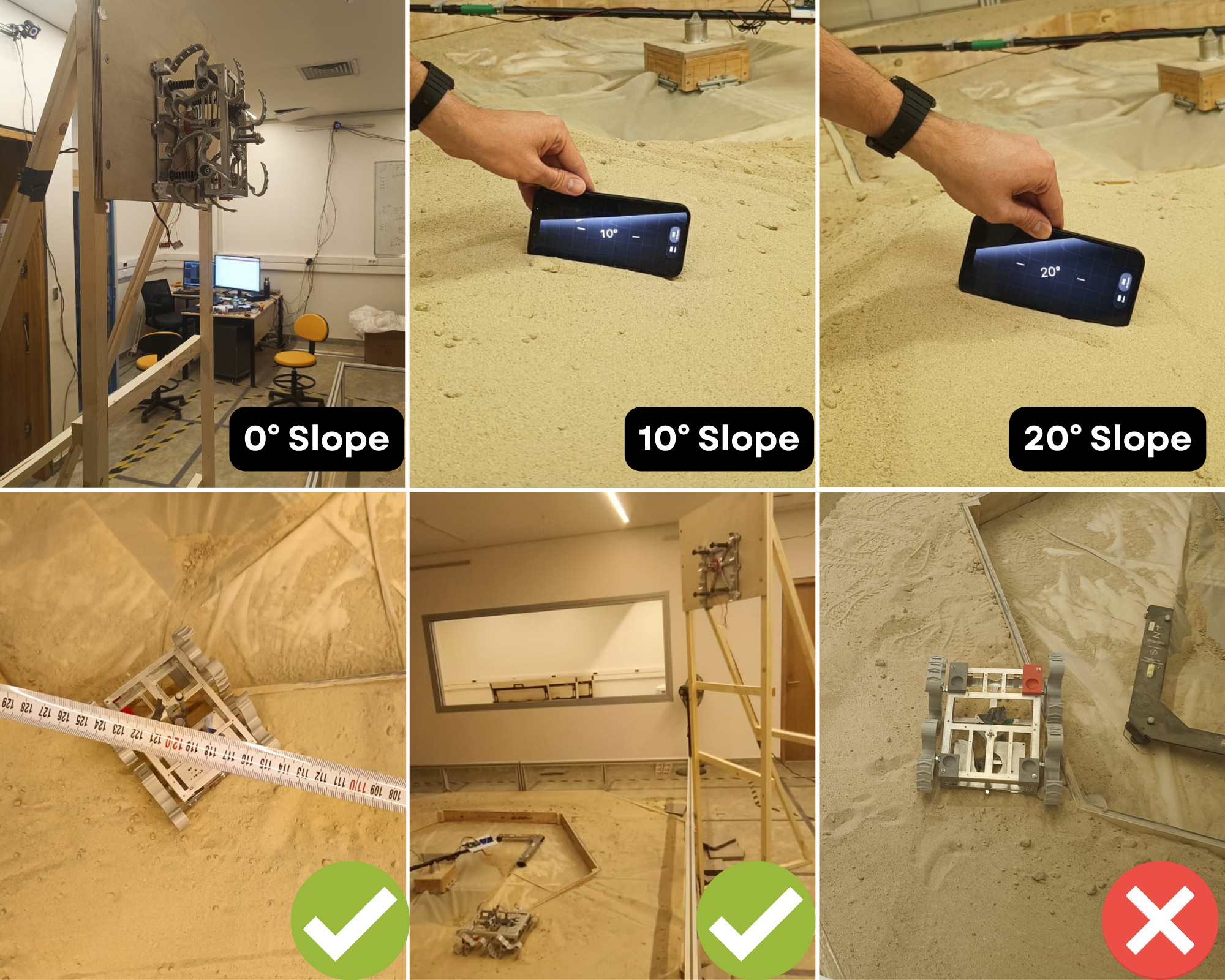}
\caption{Terrestrial test rig at three representative slope conditions used in
the campaign of Section~\ref{sec:slope_campaign}: (left to right)
$0^\circ$, $+10^\circ$ and $+20^\circ$. Steeper and clockwise ($-20^\circ$ and
beyond) conditions are reported in Table~\ref{tab:slope_matrix} but not shown
here, as they were tested with fewer repetitions after early overturns
(Section~\ref{sec:slope_campaign}).}
\label{fig:slope_scenes}
\end{figure}

Series 2 and 3 share gravity, mass and spring specification and are therefore
directly comparable; their relationship measures model fidelity alone. Series 1
differs from Series 2 only in the two parameters the laboratory cannot alter,
and is therefore the model's flight prediction. Representative terrestrial rig
configurations at three of the tested slope angles are shown in
Fig.~\ref{fig:slope_scenes}.

Each simulated case was repeated ten times, with run-to-run variation
introduced through varying initial contact friction and micro-positioning. Experimental cases were repeated ten times at $-10^\circ$, $0^\circ$ and
$+10^\circ$, and five times at the remaining inclinations, where testing was
stopped after repeated failures: each run consumes a separation nut, and
continuing to expend units on a configuration that had already failed
repeatedly was not justified.

A landing was classified as successful if the rover came to rest with all wheels
in contact with the surface and a final tilt, measured relative to the local
slope surface rather than absolute vertical, within the
$\pm 15^\circ$ stability limit; any overturn was a failure. This is a
mission-operability criterion for the settled state across the full slope range
tested, and is distinct from the $\pm 5^\circ$ in-flight attitude target of
Section~\ref{sec:mechanical_design}, which governs the separation mechanism's
kinematic behavior during free flight rather than the state after
slope-induced bounce and roll. Outcomes
are collected in Table~\ref{tab:slope_matrix}.

\begin{table}[htbp]
\centering
\caption{Separation and landing outcomes (successes/runs) over the slope
range. Lunar simulation: $5.0\,\text{kg}$, $g=1.62\,\text{m/s}^2$. Terrestrial
simulation and experiment: both $0.833\,\text{kg}$, $g=9.81\,\text{m/s}^2$.
Positive slopes are counter-clockwise, negative clockwise. Cells with $5$ runs
were truncated after repeated failure to conserve separation-nut actuations.}
\label{tab:slope_matrix}
\begin{tabular}{r ccc}
\toprule
Slope (deg) & Lunar sim. & Earth sim. & Earth exp. \\
\midrule
$-45$ & $7/10$  & $0/10$          & $0/5$ \\
$-30$ & $10/10$ & $2/10$          & $0/5$ \\
$-20$ & $10/10$ & $\mathbf{0/10}$ & $0/5$ \\
$-10$ & $10/10$ & $8/10$          & $3/10$ \\
$0$   & $10/10$ & $8/10$          & $4/10$ \\
$+10$ & $10/10$ & $9/10$          & $5/10$ \\
$+20$ & $10/10$ & $9/10$          & $\mathbf{0/5}$ \\
$+30$ & $10/10$ & $5/10$          & $0/5$ \\
$+45$ & $8/10$  & $4/10$          & $0/5$ \\
\bottomrule
\end{tabular}
\end{table}

Three features of Table~\ref{tab:slope_matrix} are noted before they are
interpreted.

The success rates are consistently ordered at every slope tested: lunar
simulation $\ge$ terrestrial simulation $\ge$ terrestrial experiment, without a
single exception across nine inclinations. This ordering is the principal
empirical result of the campaign and is examined in
Section~\ref{sec:divergence}.

Sample sizes are small and the resulting intervals are wide. Quoting
two-sided Clopper--Pearson $95\%$ intervals~\cite{clopper1934use}, the lunar $7/10$ at $-45^\circ$ is
$[35,93]\%$ and the $8/10$ at $+45^\circ$ is $[44,97]\%$; neither supports a
claim of reliable operation at $45^\circ$. Conversely, a $0/5$ outcome bounds the
success probability only at $\le 52\%$, and $0/10$ at $\le 31\%$: the
experimental failures establish that these configurations are unreliable, not
that they are impossible.

The terrestrial simulation is not monotonic in slope on the clockwise
side, returning $0/10$ at $-45^\circ$, $2/10$ at $-30^\circ$, $0/10$ at
$-20^\circ$ and $8/10$ at $-10^\circ$, so that the $-20^\circ$ case is worse
than the steeper $-30^\circ$ case. Non-monotonic outcomes are plausible in
granular landing, where the interaction between residual pitch rate, first
contact point and slope may favor recovery at one inclination and not at a
neighboring one.

\subsection{Comparative Analysis: Trajectory and Attitude}

\begin{figure}[htbp]
\centering
\includegraphics[width=0.48\textwidth]{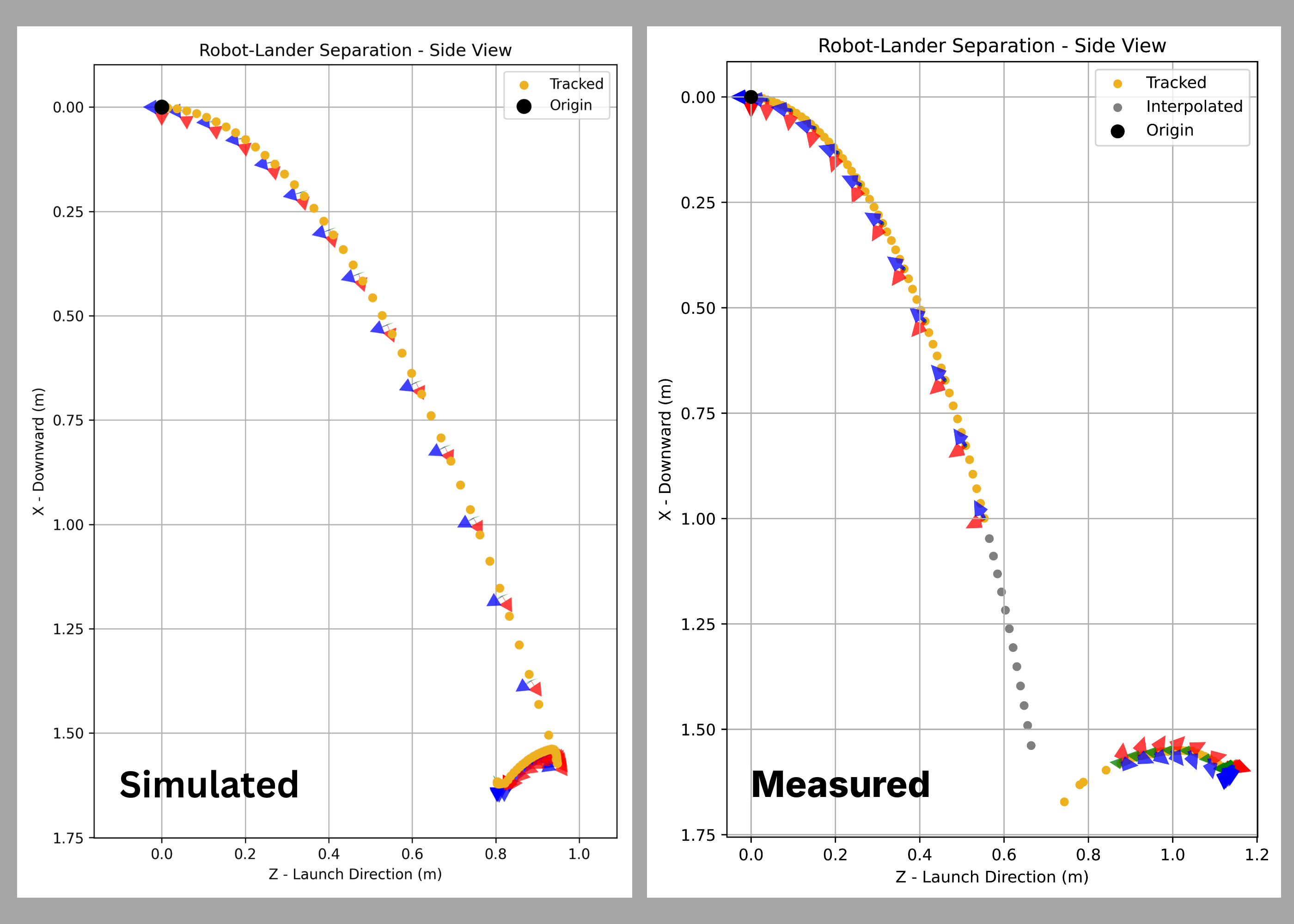}
\caption{Motion-capture trajectory of the $1/6$ mass model during separation at
$+20^\circ$ slope (right), compared with the corresponding terrestrial
simulation (left). Gray markers denote interpolated positions where tracking
data were unavailable.}
\label{fig:optitrack_trajectory+20}
\end{figure}

\begin{figure}[htbp]
\centering
\includegraphics[width=0.48\textwidth]{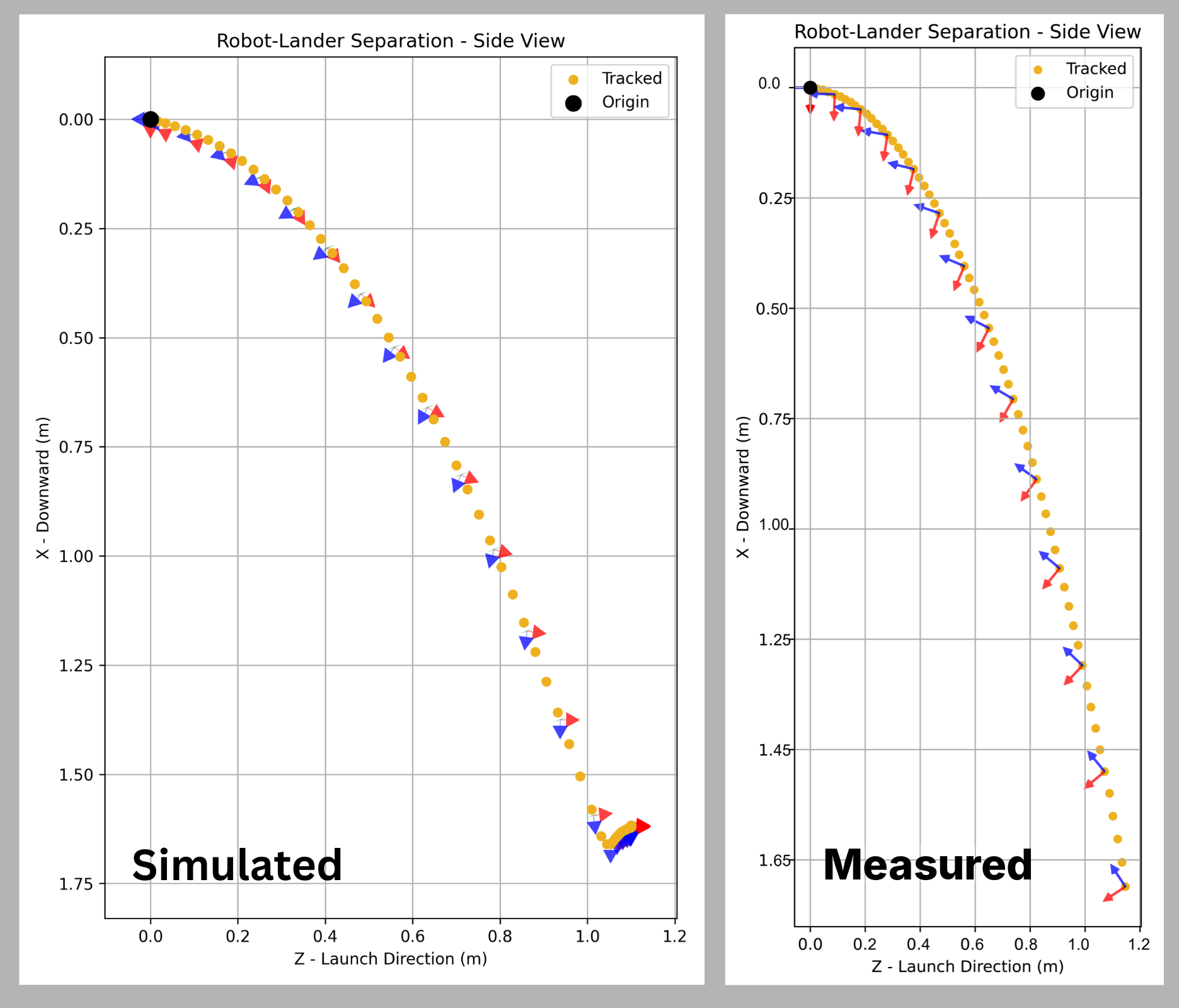}
\caption{Motion-capture trajectory of the $1/6$ mass model during separation at
$-20^\circ$ slope (right), compared with the corresponding terrestrial
simulation (left).}
\label{fig:optitrack_trajectory-20}
\end{figure}

The filtered experimental trajectories were compared with the corresponding
terrestrial simulations at two representative non-zero slope conditions,
$+20^\circ$ and $-20^\circ$; the results are shown in
Fig.~\ref{fig:optitrack_trajectory+20} and
Fig.~\ref{fig:optitrack_trajectory-20}, respectively.

At $-20^\circ$ the measured and simulated trajectories nearly coincide up to
first ground contact. At $+20^\circ$ the measured trajectory is shorter than the
simulated one: at $1\,\text{m}$ of drop the simulated center of mass is about
$0.76\,\text{m}$ downrange, against $0.56\,\text{m}$ measured, and the final part
of the measured descent is interpolated
(Fig.~\ref{fig:optitrack_trajectory+20}). The two cases also sit at opposite
ends of the outcome spectrum in Table~\ref{tab:slope_matrix}: at $+20^\circ$ the
simulation predicts $9/10$ successful landings against $0/5$ measured, the
largest divergence in landing outcome recorded in the campaign, whereas at
$-20^\circ$ simulation and experiment agree that the configuration fails to
land ($0/10$ and $0/5$, respectively). The close trajectory match at $-20^\circ$
thus coexists with a failed landing, so matching free-flight kinematics are
not, by themselves, informative about the eventual landing outcome. This
reinforces the separation argued in Section~\ref{sec:divergence} between the
flight phase and the touchdown and settling phase.

The horizontal ejection velocity, obtained from the free-flight segment of the
motion-capture data, was $1.13\,\text{m/s}$ in the $+20^\circ$ run and
approximately $1.9\,\text{m/s}$ in the $-20^\circ$ run, corresponding to
$0.46\,\text{m/s}$ and approximately $0.77\,\text{m/s}$ under lunar conditions
(Eq.~\eqref{eq:froude_velocity}); the simulation gives approximately
$1.7$--$1.9\,\text{m/s}$ in both cases. Because the free flight is completed
before ground contact, the slope cannot account for this difference; it
reflects run-to-run variation of the ejection itself, for example in the
friction of the guide housings and in the release of the separation nut,
neither of which is represented in the model. With two tracked runs, this
variation cannot be characterized statistically, and the model--experiment
difference lies within it. The corresponding translational kinetic energy,
between $\tfrac{1}{2}(0.833)(1.13)^2 \approx 0.53\,\text{J}$ and approximately
$1.5\,\text{J}$, is about $20$--$55\%$ of the
$4 \times 680.3\,\text{mJ} \approx 2.72\,\text{J}$ delivered by the spring array
over its stroke; the remainder is absorbed by J-leg deformation and structural
damping, rotation, and friction, consistent with the parasitic loss discussed
in Section~\ref{sec:analytical_optimization}. The lower measured velocity is
marginally below the $0.48\,\text{m/s}$ point-mass requirement of
Eq.~\eqref{eq:vx_req}; for a drag-free CoM trajectory it corresponds to a
horizontal travel of about $0.67\,\text{m}$. In the $+20^\circ$ run, the
horizontal travel of the center of mass measured with the motion-capture
system was about $0.74\,\text{m}$ at first ground contact and $0.84\,\text{m}$
after the first bounce, both above the
$D_{\text{clearance}} \ge 0.7\,\text{m}$ requirement. During spring expansion
($0 < t < 0.08\,\text{s}$) the flexible J-legs deform asymmetrically and
introduce a transient pitching moment; the measured peak pitch was
$+2.4^\circ$ against a predicted $+2.1^\circ$, a difference of $0.3^\circ$.
The difference is attributed to this J-leg deformation, which the rigid-body
model represents only through the lumped chassis damping. Out-of-plane rotation
remained negligible throughout the free flight, with a roll of about
$0.1^\circ$, so Figs.~\ref{fig:optitrack_trajectory+20}
and~\ref{fig:optitrack_trajectory-20} show the motion in the separation plane
only. Pitch and roll therefore stayed within the $\pm 5^\circ$ in-flight
attitude target of Section~\ref{sec:mechanical_design}.

In free flight, the model therefore agrees with the experiment to within the
run-to-run variation of the ejection, and reproduces the peak pitch during the
ejection stroke to within $0.3^\circ$. This agreement does not extend to the
landing outcome. On level
terrain, the physical model came to rest in a valid configuration in only $4$
of $10$ attempts ($40\%$, $95\%$ interval $[12,74]\%$), against $8$ of $10$ in
the corresponding simulation. Level-ground separation cannot therefore be
described as reliably achieving a stable wheels-first touchdown, and is not so
described here. The free-flight phase is captured by the model within the
scatter of the ejection; the touchdown and settling phase is not.

\subsection{Two Distinct Divergences and Their Origins}
\label{sec:divergence}

The ordering noted above decomposes into two independent gaps with different
causes and different implications.

\subsubsection{Gap A: Lunar versus Terrestrial Simulation}
\begin{figure}[htbp]
\centering
\includegraphics[width=0.48\textwidth]{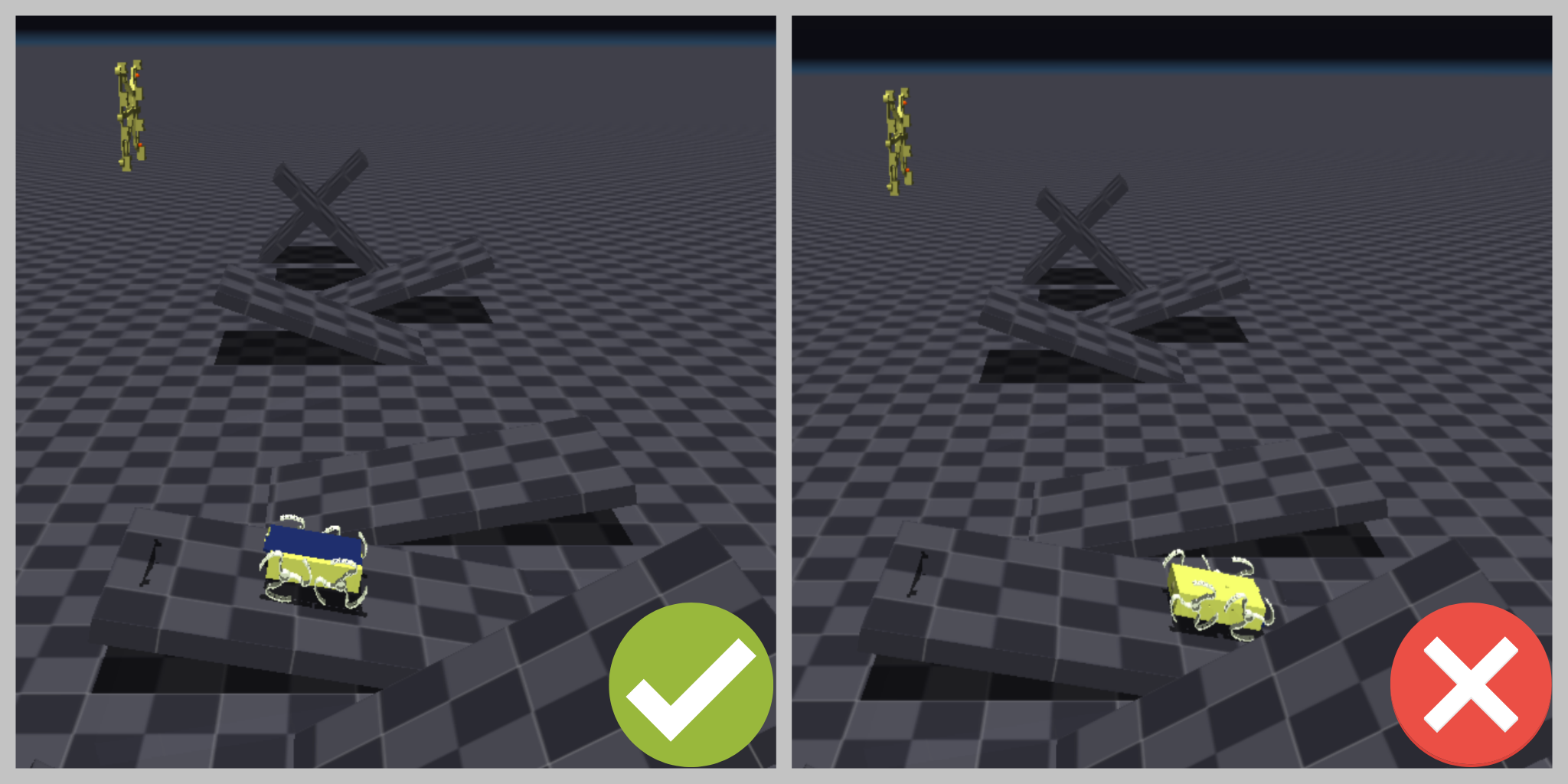}
\caption{Simulated landing outcomes on a $-20^\circ$ slope. Left: lunar
conditions, full flight mass, successful wheels-first landing. Right:
terrestrial conditions, $1/6$ mass model, overturned. The free-flight
trajectories are geometrically identical; the divergence occurs entirely during
touchdown and settling.}
\label{fig:landing_comparison}
\end{figure}

Both use the same solver, model, spring specification and contact parameters
and differ only in gravity and mass. The lunar case succeeds in $10/10$ runs for
$|{\rm slope}| \le 30^\circ$ and falls to $7/10$ and $8/10$ at $-45^\circ$ and
$+45^\circ$; the terrestrial case fails entirely at $-45^\circ$ and $-20^\circ$, nearly so
at $-30^\circ$ ($2/10$), and is degraded elsewhere
(Fig.~\ref{fig:landing_comparison}). The rover arrives with the same kinetic
energy and the same attitude in both cases, but at about $2.46$ times the
velocity and with one sixth of the mass (Table~\ref{tab:scaling}). Because the
contact parameters are not scaled with gravity, the contact model does not
preserve Froude similarity during touchdown: the faster impact of the lighter
body produces a different contact response, and this response, rather than the
free flight, determines whether the residual rotation carries the center of
mass beyond the support polygon. Gap A therefore reflects the gravity scaling
combined with a contact model that is not scaled with it; it is not a property
of the separation mechanism. In the physical experiment, the approximately
six-fold higher frictional bearing capacity of the simulant under terrestrial
gravity (Section~\ref{sec:gravity_scaling}) is an additional difference from
lunar conditions that the simulations do not include.

\subsubsection{Gap B: Terrestrial Simulation versus Terrestrial Experiment}
These share gravity, mass, springs and slope, so any difference is model
fidelity. The simulation never under-predicts success, and over-predicts it at every
inclination where the two differ: by $20$ percentage points at $-30^\circ$, $50$
at $-10^\circ$, $40$ at $0^\circ$, $40$ at $+10^\circ$, $50$ at $+30^\circ$ and
$40$ at $+45^\circ$, and by $90$ percentage points at $+20^\circ$, where the
model returns $9/10$ against $0/5$ measured. Gap B is not
attributable to gravity scaling and is not explained by it.

Candidate contributions, in order of expected significance, are: representation
of the compliant J-legs as part of a rigid body, with their energy absorption
lumped into chassis damping, so that leg deflection and its capacity to catch
or dig at touchdown are absent from the model; the convex-hull
treatment of the collision geometry, which fills concavities and enlarges the
effective support polygon; uncontrolled variation in the compaction state and
surface preparation of the simulant between runs; and asymmetry in the release
of the separation nut, idealized as instantaneous and symmetric in the model. Discriminating
between these requires instrumented repeat testing with controlled simulant
preparation, and is identified as required future work.

The $+20^\circ$ discrepancy is singled out as both the largest and the least
consistent with its neighbors: the experimental result falls from $5/10$ at
$+10^\circ$ to $0/5$ at $+20^\circ$ while the simulation is essentially
unchanged across the same interval. It is reported as measured, but no
conclusion is rested upon it.

\subsubsection{Implication for the Flight Prediction}
The two gaps act in opposite directions and together bracket the expected flight
behavior rather than establishing it. The terrestrial experiment is
conservative relative to flight, because lunar regolith absorbs more of the
arrival energy than the terrestrially loaded simulant. The simulation is
optimistic relative to a physical article, as Gap B demonstrates under matched
conditions, and the lunar simulation inherits that optimism because it uses the
same model. The lunar predictions of Table~\ref{tab:slope_matrix} should
therefore be read as an upper bound on flight performance and the terrestrial
results as a lower bound, with the expectation that flight behavior lies
between them. No single figure in this campaign is a measurement of lunar
landing reliability, and none is presented as one.

The failure is asymmetric in slope sign because the separation event imparts a
net pitch rate in a single sense, arising from the offset of the resultant spring force vector relative to the center of mass. A clockwise slope adds to this residual rotation
while a counter-clockwise slope opposes it, which is consistent with the
poorer clockwise performance in the terrestrial simulation and, marginally, at
$\pm 45^\circ$ in the lunar simulation.

\subsection{Scope of Validation, Design Consequences and Future Work}
 
What has been established is that the numerical model reproduces the free-flight
trajectory to within the run-to-run variation of the measured ejection velocity,
and the peak pitch during the ejection stroke to within $0.3^\circ$, against
experiments that were not used to calibrate it. This supports the spring model,
the release geometry and the free-flight integration; characterizing the
ejection velocity statistically will require more tracked runs.
 
What remains unclear is whether the model predicts landing outcomes. Gap B
shows that it does not, under the one set of conditions where it can be checked
directly. The average slope of the pre-selected landing zone is specified as
less than $8^\circ$~\cite{bice2025devspec}, which lies below the
inclination range at which slope-driven overturns appear in the simulated
series and the physical article. That comparison should not, however, be read as margin. Within
$\pm10^\circ$ the measured success rates are $3/10$, $4/10$ and $5/10$ at
$-10^\circ$, $0^\circ$ and $+10^\circ$ respectively; the Clopper--Pearson
intervals overlap throughout and no slope dependence is resolvable over that
range. Pooling the three gives $12/30$ against $0/30$ for
$|{\rm slope}| \ge 20^\circ$, a difference that is significant (two-sided Fisher
exact test, $p \approx 1\times10^{-4}$) and establishes that slope governs the outcome beyond
$20^\circ$. Inside the site envelope it does not. The binding limitation there
is the baseline landing reliability of approximately $40\%$ observed on level
ground, which a shallow site does nothing to relieve.
 
Two further cautions attach to the site figure itself. A zone-average slope is
derived at baselines of tens of meters, whereas the stability of a
$0.35\,\text{m}$-scale vehicle is set by the local surface beneath its own
footprint; boulders, small craters and their ejecta produce local inclinations
substantially steeper than the regional mean. And the $\pm45^\circ$ envelope is
a prediction of the model whose optimistic bias has just been quantified, so it
cannot carry a margin argument on its own. The defensible statement is that
slope is not expected to be the dominant risk at this site, and that the
dominant risk is the landing reliability of the separation event itself,
irrespective of terrain.
 
Two design consequences follow. First, the residual pitch rate imparted at
separation is the main controllable parameter expected to affect the baseline
landing reliability; balancing the asymmetric spring array moments about the
center of mass (Section~\ref{sec:mechanical_design}) is the most direct route to
mitigating this risk and is the recommended first action.
 
Second, the results confirm that the second layer of the deployment strategy
set out in Section~\ref{sec:introduction}---the self-righting recovery
mechanism, in which coordinated J-leg maneuvers combined with deployment of the
top-mounted solar panel act as levers against the regolith to restore an
inverted chassis---is a requirement rather than a contingency. The physical
article overturned at inclinations as low as $20^\circ$ and failed to achieve a
valid configuration in the majority of level-ground attempts, so a deployment
architecture without recovery cannot be considered flight-worthy on the evidence
presented here.
 
Quantifying that recovery capability is beyond the scope of the present work,
which addresses the separation event itself, and is the principal item of
ongoing development. Establishing it will require the reaction torque available
from the coupled leg--panel kinematics against a regolith of the relevant
bearing capacity; the actuation and energy budget of the maneuver under
flight power constraints; the range of inverted attitudes and slopes from which
recovery is feasible; and the dust-tolerance of the deploying panel hinge after
a regolith-contacting maneuver. Until those are established, the landing
reliability reported in Table~\ref{tab:slope_matrix} should be read as the
as-landed figure, with no credit taken for recovery.
 
Three further lines of work follow from the gaps identified above. The
decomposition of Gap B requires instrumented repeat testing with controlled
simulant preparation and compaction measurement, together with a compliant
representation of the J-legs in place of the present rigid-body treatment.
The non-monotonic clockwise response of the terrestrial simulation between
$-30^\circ$ and $-20^\circ$ warrants a finer slope sweep to determine whether it
is a physical effect or an artifact of the contact model. Finally, validating the touchdown
dynamics that weight-matched scaling cannot reproduce (the lower block of
Table~\ref{tab:scaling}) would require a genuine reduced-gravity facility, whether parabolic flight or an active offload rig applying a constant
upward force to the full-mass article; only such a test can validate the
landing phase rather than the flight phase. In addition, the lander
specification allows the lander to rest up to $15^\circ$ from the local
horizontal after landing~\cite{bice2025devspec}, which tilts the ejection
direction of the laterally mounted rovers; the effect of this launch-angle
variation on clearance and landing attitude has not been tested and is left to
future work.

\section{Conclusion}
\label{sec:conclusion}

In this study, a lightweight mechanical separation system was developed and
tested for a $5.0\,\text{kg}$ class lunar micro-rover. Over six design
generations, the mechanical, structural, and kinematic problems of stowing
and deploying a lightweight rover were identified in test and addressed in
redesign.

The main results of this work are summarized as follows:
\begin{enumerate}
    \item \textbf{Multidisciplinary Design Evolution:} The separation
    system evolved from a dual shape memory alloy (SMA) release-nut
    architecture, sensitive to actuator synchronization, into a
    single-actuator hold-down and release mechanism utilizing a central fracture-type
    separation nut~\cite{sfr_datasheet}, coupled with an asymmetric
    four-spring ejector array. This transition resolved the limitation that the release-only micro SMA
    release nut~\cite{nD3RN_datasheet} was not sufficiently repeatable under
    the installation torques and bending moments of the stiffer spring array
    during repeated ground testing.

    \item \textbf{Parametric and Topological Optimization:} A two-stage
    parametric spring optimization balanced the energy stored across the
    four-spring array and increased the spring rate while keeping the energy
    delivered per spring over the separation stroke constant at
    $680.3\,\text{mJ}$ and the spring force at the release point at
    $\approx 12.1\,\text{N}$ (versus $9.8\,\text{N}$ for the shorter Proposed
    design); together with the re-engineered guides, this eliminated the
    jamming observed in vibration testing. A Solid Isotropic Material with Penalization (SIMP)
    topology optimization then redesigned the flat ground plate as a ribbed
    plate ($1.5\,\text{mm}$ webs, ribs up to $8.0\,\text{mm}$), raising the
    stowed assembly's first natural frequency from a measured
    $34\,\text{Hz}$ to $84\,\text{Hz}$ ($+147\%$), above the
    $70\,\text{Hz}$ lander requirement, within the $1.6\,\text{kg}$ mass
    budget.

    \item \textbf{Environmental Testing:} The
    sixth-generation separation hardware was tested at the satellite testing
    facilities of T\"{U}B\.{I}TAK UZAY. Set-up checks and shock calibration were
    performed with a $5.0\,\text{kg}$ dummy model, and the test runs with a
    Structural Model (SM). Under sinusoidal and random vibration in all
    three orthogonal axes at acceptance level (up to $6\,\text{g}$
    sinusoidal and $9.8\,\text{g}_{\text{rms}}$ random), the assembly showed no structural
    damage, with pre- and post-test
    resonance search sweeps ($5$--$2200\,\text{Hz}$ at $0.5\,\text{g}$)
    showing frequency shifts within the $\le 5\%$ pass criterion. In
    pneumatic shock testing according to ECSS-E-ST-10-03C, at a specified
    SRS reaching $2200\,\text{g}$ over $1500$--$4000\,\text{Hz}$, no
    plastic deformation or premature release occurred.

    \item \textbf{Kinematic and Trajectory Validation:} The separation
    trajectory was evaluated in the laboratory using a weight-matched
    $1/6$ mass model ($0.833\,\text{kg}$) deployed into a regolith simulant
    sandbox. 6-DOF tracking with an eight-camera, $240\,\text{Hz}$
    infrared motion-capture system showed that the MuJoCo multi-body
    simulation reproduces the free flight to within the run-to-run
    variation of the ejection. The horizontal ejection velocity, referred
    to lunar conditions, was $0.46\,\text{m/s}$ and approximately
    $0.77\,\text{m/s}$ in the two tracked runs, against the
    $0.48\,\text{m/s}$ requirement, and the transient pitch angle during
    the ejection stroke reached a measured peak of $+2.4^\circ$ against a
    predicted $+2.1^\circ$, with a roll of about $0.1^\circ$, within the
    $\pm 5^\circ$ in-flight attitude target. The horizontal travel of the
    center of mass was about $0.74\,\text{m}$ at first ground contact and
    $0.84\,\text{m}$ after the first bounce, exceeding the $0.7\,\text{m}$
    requirement. Only
    about $20$--$55\%$ of the energy delivered by the springs appeared as
    translational kinetic energy, confirming substantial parasitic losses
    in the J-legs and chassis. These results support the spring model and
    release geometry in free flight; they do not by themselves establish
    landing-outcome reliability, which was characterized separately across
    a range of terrain slopes and is discussed below.
\end{enumerate}

The combination of the SIMP-optimized ribbed plate, low-stiction guides, and
the weight-matched mass-model test approach provides a low-cost framework for
developing and testing lightweight planetary deployment mechanisms, provided
that its limits, in particular for the touchdown phase, are respected.

The slope campaign showed that baseline landing reliability (only $4/10$
upright landings of the physical model on level ground), rather than terrain
slope, is the dominant risk within the site's expected topography, so the self-righting recovery maneuver
described in this work is a required capability rather than a contingency.
Quantifying it --- its reaction-torque and actuation-energy budget, the
range of recoverable attitudes and slopes, and the dust-tolerance of the
panel hinge after regolith contact --- is accordingly the principal item of
ongoing development.

Future work will also extend the multi-body dynamic
models by coupling Discrete Element Method (DEM) solvers to simulate
cohesive wheel-leg--regolith interaction forces and slip during
touchdown and settling. The estimation of the rover's state after landing,
which determines whether the self-righting maneuver is needed, is the subject
of a follow-up study. Additionally, the separation mechanism will be
subjected to thermal-vacuum (TVAC) testing campaigns to evaluate the
thermal fatigue and coefficient-of-friction stability of the dry
$\text{MoS}_2$ lubricant film on the hard-anodized aluminum (Al 7075-T6)
guide bushings across the extreme cryogenic and high-temperature cycles of
the lunar night and day, toward flight readiness.

\bmsubsection*{Author Contributions}
\textbf{Ahmet Mustafa Kangal:} conceptualization, methodology, software, formal analysis, investigation, validation, visualization, writing -- original draft, writing -- review and editing.
\textbf{O\u{g}uzhan \c{S}ahan:} investigation (motion-capture separation tests), data curation.
\textbf{Teoman Efe Aycan:} methodology and formal analysis (topology optimization).
\textbf{G\"{o}rkem Egemen G\"{u}lo\u{g}lu:} supervision, writing -- review and editing.
\textbf{Halil Ersin S\"{o}ken:} supervision, project administration, writing -- review and editing.
\textbf{Mustafa Mert Ankaral{\i}:} supervision, writing -- review and editing.
\textbf{Uluç Saranl{\i}:} supervision.
\textbf{Ali Emre Turgut:} supervision.

\bmsubsection*{Acknowledgments}
The authors thank the entire CHERI project team for their contributions to the design, integration, and testing of the rovers, and the METU Robotics and Artificial Intelligence Technologies Application and Research Center (ROMER) for providing laboratory facilities, including the motion-capture test arena. The authors also thank T\"{U}B\.{I}TAK UZAY for their support during the vibration and shock test campaigns, and Karakamlar INC. for their support with the center-of-mass and moment-of-inertia measurements of the CHERI rovers.

\bmsubsection*{Financial Disclosure}
This study is supported in part by the ADEP project with grant number ADEP-313-2025-11656.

\bmsubsection*{Conflicts of Interest}
The authors declare no conflicts of interest.

\bmsubsection*{Data Availability Statement}
The data that support the findings of this study are available from the corresponding author upon reasonable request.

\bibliography{references}

\end{document}